\documentclass{article}

\usepackage[preprint]{neurips_2026}   %
\makeatletter\renewcommand{\@noticestring}{}\makeatother   %

\usepackage[utf8]{inputenc}
\usepackage[T1]{fontenc}
\usepackage{hyperref}
\usepackage{url}
\usepackage{booktabs}
\usepackage{placeins}   %
\usepackage{array}
\usepackage{float}
\usepackage{amsmath, amssymb, amsfonts, amsthm}
\usepackage{nicefrac}
\usepackage{microtype}
\usepackage{xcolor}
\usepackage{graphicx}
\usepackage{subcaption}
\usepackage{pgfplots}
\usepackage{twemojis}   %

\usepackage[capitalize, noabbrev]{cleveref}

\graphicspath{{figures/draft/}{figures/}}

\newif\ifnotes \notestrue

\newcommand{\ours}{RumiFold}
\newcommand{\foldonly}{RumiFold-Struct}   %
\newcommand{\esmc}{ESM-C}
\newcommand{\lddt}{lDDT}
\newcommand{\fmax}{F$_{\max}$}

\pgfplotsset{compat=1.18}
\usetikzlibrary{patterns}

\definecolor{rumiGreen}{HTML}{22754A}   %
\definecolor{rumiAmber}{HTML}{E09A3A}   %
\definecolor{rumiBlue}{HTML}{5D6DB8}    %
\definecolor{rumiBrown}{HTML}{A0522D}   %
\definecolor{rumiMint}{HTML}{9CC9AC}    %
\definecolor{rumiGrey}{HTML}{8E9399}    %
\definecolor{rumiGreyL}{HTML}{CFD3D7}   %
\definecolor{rumiInk}{HTML}{222222}     %
\colorlet{cOurs}{rumiGreen}  \colorlet{cCtrl}{rumiMint}
\colorlet{cBaseA}{rumiAmber} \colorlet{cBaseB}{rumiBlue} \colorlet{cBaseC}{rumiBrown}
\colorlet{cGrey}{rumiGrey}   \colorlet{cGreyL}{rumiGreyL}

\pgfplotsset{colormap={rumi}{color(0)=(white) color(0.5)=(rumiMint) color(1)=(rumiGreen!75!black)}}

\pgfplotsset{
  paper/.style={
    axis lines*=left, ymajorgrids, grid style={gray!18},
    tick align=outside, major tick length=2pt,
    tick label style={font=\scriptsize}, label style={font=\scriptsize},
    title style={font=\scriptsize\bfseries, yshift=-4pt},
    legend style={font=\scriptsize, draw=none, fill=none, legend cell align=left,
                  /tikz/every even column/.append style={column sep=6pt}},
    legend image code/.code={\draw[##1, draw=none] (0cm,-0.08cm) rectangle (0.22cm,0.1cm);},
  },
  vals/.style={
    nodes near coords,
    every node near coord/.append style={font=\tiny, rotate=90, anchor=west, inner sep=1.5pt,
      text=rumiInk, /pgf/number format/.cd, fixed, fixed zerofill, precision=#1},
  },
  reference bar/.style={fill=rumiGreyL, draw=rumiGrey,
    postaction={pattern=north east lines, pattern color=rumiGrey}},
}

\usetikzlibrary{arrows.meta, positioning, fit, calc, backgrounds, shapes.geometric, decorations.pathreplacing}
\tikzset{
  sch/.style={font=\scriptsize, text=rumiInk, >={Stealth[length=4.5pt,width=4pt]}, line cap=round},
  box/.style={draw=rumiGrey, fill=white, rounded corners=2.5pt, line width=0.5pt,
              align=center, inner sep=3.5pt, minimum height=6.5mm},
  rumi/.style={box, draw=rumiGreen, fill=rumiGreen!12, line width=0.7pt},
  rumidark/.style={box, draw=rumiGreen, fill=rumiGreen, text=white, line width=0.7pt},
  seq/.style={box, draw=rumiAmber, fill=rumiAmber!16},
  ctx/.style={box, draw=rumiGrey, fill=rumiGreyL!35},
  txt/.style={box, draw=rumiBlue, fill=rumiBlue!10},
  flow/.style={->, line width=0.7pt, draw=rumiInk!75},
  foldflow/.style={->, line width=0.9pt, draw=rumiGreen},
  seqflow/.style={->, line width=0.9pt, draw=rumiAmber!90!black},
  faint/.style={->, line width=0.6pt, draw=rumiGrey, densely dashed},
  lab/.style={font=\tiny, text=rumiInk!70, align=center},
  head/.style={font=\scriptsize\bfseries, text=rumiInk},
  stage/.style={font=\scriptsize\bfseries, text=white, fill=#1, rounded corners=2pt, inner sep=2.5pt},
}

\title{\ours{}: Teaching an LLM to Natively Fold Proteins and Reason About Their Function}

\author{%
  Kevin Lu \\
  Northeastern University \\
  \texttt{lu.kev@northeastern.edu}
  \And
  David Bau\thanks{Co-advisors and co-corresponding authors.} \\
  Northeastern University \\
  \texttt{d.bau@northeastern.edu}
  \And
  Wengong Jin\footnotemark[1] \\
  Northeastern University \\
  \texttt{w.jin@northeastern.edu}
}

\begin{document}

\maketitle

\begin{abstract}
Can we teach a general-purpose language model to fold proteins? And if so, what does learning to fold enable beyond structure prediction itself? We introduce RumiFold, a pretrained chat model that learns to predict protein structures directly from its own hidden states. RumiFold processes proteins through repeated folding passes, then uses these folded representations to answer questions in natural language. It achieves folding accuracy comparable to dedicated single-representation models such as SimpleFold, alongside state-of-the-art performance on Gene Ontology protein function prediction and free-text function description. We find that folding pretraining organizes the model's hidden states around protein function, creating an internal per-residue workspace that is biologically grounded. Causal interventions show that this folded workspace links structure and language when RumiFold predicts function, providing rich representations for biomolecular reasoning. We see RumiFold as a step toward \emph{native biomolecular intelligence}: where protein-based computations and language-based reasoning become integrated capabilities of general-purpose language models. Code, data, and models are available at: \href{https://rumifold.baulab.info}{\texttt{rumifold.baulab.info}}.

\end{abstract}

\section{Introduction}
\label{sec:intro}

Can an LLM learn to fold proteins? And if it can, what does it gain beyond the ability to predict structure? An LLM can already call AlphaFold3~\citep{abramson2024accurate} to obtain a protein structure, but doing so does not teach it how to fold. By learning to perform this biomolecular computation itself, a language model could develop an internal workspace that supports richer reasoning about protein structure and function.

Current efforts to connect general-purpose language models with proteins largely leave this opportunity open by keeping biomolecular computation outside the model's own representations. One path relies on tool-calling and agentic workflows around external predictors and biological software~\citep{huang2025biomni,ghafarollahi2024protagents,swanson2025virtual}, where biological computation happens inside the tool, forcing the language model to reason over lossy text summaries rather than more expressive internal representations. Another path bridges sequence or structure encoders to text LLMs~\citep{fei2025prot2text,fallahpour2026bioreason,liu2024evollama,wang2025prot2chat}, relying on static features or precomputed predictions to generate descriptions rather than building an internal representation of the protein for reasoning. What is missing is a general-purpose chat model that folds proteins within its own hidden states and reasons from the result, in place of reading a structure or a prediction produced by another model.

\begin{figure}[t]
  \centering
  \includegraphics[width=\linewidth]{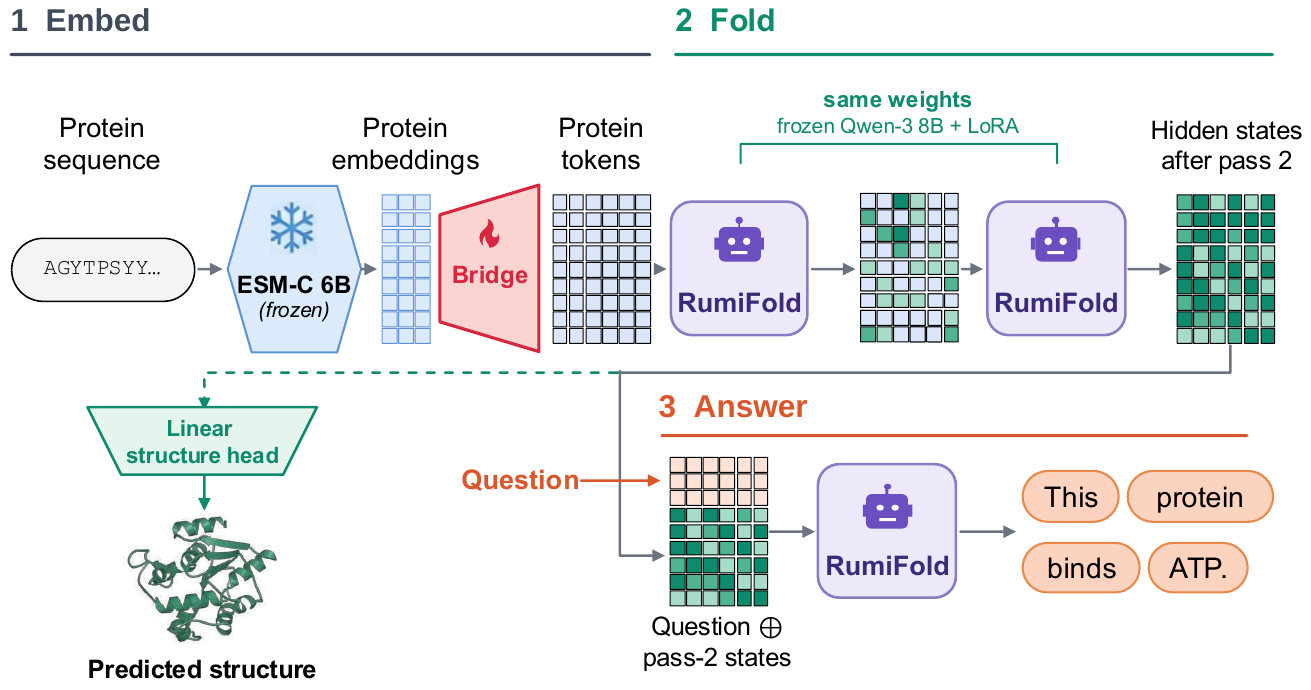}

\caption{\textbf{\ours{} embeds a protein, folds it, and then answers from what it learned.}
  \emph{(1)~Embed:} A frozen \esmc{} encoder embeds each residue, and a trained bridge maps the embeddings into the language model's hidden space.
  \emph{(2)~Fold:} The language model (frozen Qwen3-8B with LoRA adapters) processes the protein in two folding passes, recycling the hidden states from the first pass into the second to refine its representation of protein structure. A linear structure head predicts backbone coordinates from the resulting hidden states.
  \emph{(3)~Answer:} The same language model attends to these folding representations when answering questions in natural language.}
  \label{fig:overview}
  \vspace{-6pt}
\end{figure}

We introduce RumiFold, a general-purpose language model trained to predict protein structures from its own hidden states. We fine-tune Qwen3-8B with LoRA~\citep{hu2021lora}, with a frozen ESM-C encoder~\citep{esmc2024} supplying an input embedding for each residue. Everything after is done by the language model's own layers: it processes the protein through repeated folding passes, then attends to the resulting representations when answering questions in natural language (\cref{fig:overview}).

\paragraph{Contributions: what did we find?}
\begin{itemize}
  \setlength{\itemsep}{1pt}\setlength{\parskip}{0pt}\setlength{\topsep}{2pt}
  \item \textbf{A general-purpose language model can learn to fold proteins.} With its weights frozen and only
        low-rank adapters and a linear output head trained, Qwen3-8B folds proteins as
        well as small dedicated single-representation models (FoldBench monomers: RumiFold TM-score 0.840 vs SimpleFold 700M 0.832) (\cref{sec:fold}).
  \item \textbf{Learning to fold provides representations useful for understanding biological function.} Folding pretraining reorganizes the model's hidden states around protein structure and function. As a result, RumiFold achieves state-of-the-art performance on Gene Ontology protein function prediction and free-text function description, and it outperforms both a matched model trained without folding and one that reads ESMFold2's hidden states instead (RumiFold wF1 0.683 vs GO-GPT v2 0.666; RumiFold BLEU-4 41.0 vs Prot2Text-V2 39.0) (\cref{sec:function}).
  \item \textbf{The model uses an internal workspace of representations it develops through folding.} RumiFold's folding passes construct a per-residue workspace that the language model subsequently uses to answer questions. Through causal interventions, we show that its function predictions depend on this workspace and that the underlying representations capture functionally meaningful features and residues (such as the recently characterized functional residue H130 in human TMEM68) (\cref{sec:mechanism}).
\end{itemize}

\paragraph{Toward native biomolecular intelligence.} Language models became broadly capable by working through an enormous range of problems themselves: predicting, explaining, translating and solving, until reasoning and planning were ingrained in the model. Becoming capable at biology may require the same approach: letting the model practice scientific skills until it learns the computation itself and can reason with what it has learned. We see RumiFold as a step towards \emph{native biomolecular intelligence}, where protein computations and language-based reasoning become integrated capabilities of general-purpose language models.

\section{Related Work}
\label{sec:related}

\paragraph{Protein encoders paired with LLMs.}
Most recent function predictors turn the output of a protein encoder into tokens an LLM can read
and let the LLM write the answer. GO-GPT decodes GO terms from ESM-2 features, and
BioReason-Pro \citep{fallahpour2026bioreason}, which introduced it, adds GO, InterPro and
interaction context to a reasoning LLM; Prot2Text-V2 \citep{fei2025prot2text}, Evolla
\citep{zhou2025decoding}, BioMedGPT \citep{luo2023biomedgpt}, EvoLlama \citep{liu2024evollama} and
Prot2Chat \citep{wang2025prot2chat} connect sequence or structure encoders to Llama-family models.
A second line presents the sequence as plain text
\citep{fang2023mol,wang2023instructprotein,shen2024toursynbio}. \ours{} differs in one key aspect: the
protein is first worked over by the language model itself, in a pass trained to fold it, before
the model is asked anything.

\paragraph{Do these models use the protein?}
\citet{wu2025rethinking} show that retrieval of similar training proteins rivals or beats trained
protein LLMs on text-based benchmarks, and \citet{fang2026when} show by input perturbation and
evidence-conflict tests that some models rely on accompanying text rather than on the biological
input. We adopt both lessons: every function result is stratified by sequence identity or novelty from the
training set, and we test for interpretable usage of the protein rather than relying only on accuracy numbers.

\paragraph{Structure prediction.}
Accurate protein structure prediction arrived with AlphaFold2 \citep{jumper2021highly}, which
reads a multiple sequence alignment and refines a pairwise representation of the protein over many
rounds. Two simplifications followed. Protein language models were used as BERT-style encoders that also capture the evolutionary information supplied by an MSA: ESMFold \citep{lin2023evolutionary} attaches a folding trunk, still with
pair representations, to ESM-2 and predicts structure from one sequence. A more recent group also
drops the pair representation and predicts structure from per-residue sequence representations alone, among them
ESM3 \citep{hayes2025simulating}, DPLM-2 \citep{wang2024dplm} and SimpleFold
\citep{wang2025simplefold}. \ours{} belongs to this last group, with one difference: its trunk is a
general-purpose language model, not a network trained from scratch on proteins.

\section{\ours{}}
\label{sec:method}

\subsection{Architecture (\texorpdfstring{\cref{fig:overview}}{Figure 1})}
\label{sec:arch}

\paragraph{Input.}
Each input protein chain is embedded by a frozen \esmc{}-6B encoder \citep{esmc2024}; we take the final-layer
state $\mathbf{e}_i \in \mathbb{R}^{2560}$ of each residue $i$. A bridge maps it into Qwen's
embedding space, giving the vector $\mathbf{h}^0_i \in \mathbb{R}^{d}$ that the trunk receives
in place of a token embedding at position $i$:
\begin{equation}
  \mathbf{h}^0_i = \lambda\,\mathrm{LayerNorm}\big(W\,\mathrm{LayerNorm}(\mathbf{e}_i) + \mathbf{b}\big),
  \qquad
  \lambda = \sqrt{\frac{1}{|V|\,d}\sum_{v=1}^{|V|}\big\lVert \mathbf{E}_v \big\rVert_2^2}\,,
  \label{eq:input}
\end{equation}
where $W \in \mathbb{R}^{d \times 2560}$ and $\mathbf{b}$ are learned, $d = 4{,}096$ is Qwen's hidden
width and $\mathbf{E}_v$ is Qwen's embedding of vocabulary token $v$. The constant $\lambda$ is the
root-mean-square entry of Qwen's token-embedding matrix, so residues enter the trunk at the scale
of ordinary tokens. \ours{} accepts one or two chains. Other architectural details are in
\cref{app:archdetails}.

\paragraph{Trunk.}
The trunk is Qwen3-8B \citep{yang2025qwen3}. All 36 blocks are frozen, and LoRA adapters of rank
256 are placed on the query, key, value and output projections and the three MLP projections of
every block (252 projections) for a total of about 700M parameters.

\paragraph{Three passes, two attention masks.}
\ours{} runs its trunk three times on a protein, with two different attention masks
(\cref{fig:masks} in \cref{app:masks}). The same LoRA weights serve all three passes.

\emph{Folding passes.} The first two passes take the protein alone, with the causal mask removed:
every residue attends to every residue of every chain. Let $\mathbf{s}^{(1)}_i$ be the final hidden state of residue $i$ after the first pass. The
second pass receives
\begin{equation}
  \mathbf{h}^{\mathrm{in}}_i = \mathbf{h}^0_i + \lambda\,\tanh(g)\,
     \mathrm{LayerNorm}\big(\mathbf{s}^{(1)}_i\big),
  \label{eq:recycle}
\end{equation}
as in the recycling of AlphaFold2 \citep{jumper2021highly}, where $\mathbf{h}^0_i$ and $\lambda$ are those of \cref{eq:input} and $g$ is a single
learned scalar initialised to zero, so that training starts from a model with no recycling. The
structure is read from the second pass.

\emph{Question pass.} The third pass runs causally over the text, as a language model normally
does, with the protein as a prefix it can attend to but not change. Let $\mathbf{k}^{\ell}_j$ and
$\mathbf{v}^{\ell}_j$ be the key and value that layer $\ell$ computed for residue $j$ during the
second folding pass. At layer $\ell$, the attention output for text token $t$ with query
$\mathbf{q}^{\ell}_t$ is
\begin{equation}
  \mathbf{o}^{\ell}_t = \sum_{j=1}^{L} \alpha^{\ell}_{tj}\,\mathbf{v}^{\ell}_j
     + \sum_{u \le t} \alpha^{\ell}_{tu}\,\tilde{\mathbf{v}}^{\ell}_u,
  \qquad
  \alpha^{\ell}_{t\,\cdot} = \mathrm{softmax}\big(\mathbf{q}^{\ell}_t \cdot
     [\,\mathbf{k}^{\ell}_{1:L},\ \tilde{\mathbf{k}}^{\ell}_{1:t}\,]\big),
  \label{eq:question}
\end{equation}
where tildes mark the keys and values of the text itself. This is an ordinary key--value cache,
with two differences from the one a language model builds for its own prompt: it was computed
under bidirectional attention, so each residue's entry already reflects the whole protein, and it
holds all 36 layers of the folded state rather than a single input embedding per residue. Text
positions continue directly after the protein's. Nothing flows the other way: the folding passes
never attend to the question, so the predicted structure cannot depend on what is asked.

\paragraph{From hidden states to a structure.}
A protein backbone is a chain of residues, each contributing three atoms (N, C$\alpha$, C) in a
nearly fixed local arrangement. Predicting the backbone therefore amounts to deciding, for every
residue, \emph{where} it sits and \emph{how it is oriented}: a rigid frame
$T_i = (R_i, \mathbf{t}_i)$, a rotation and a position. A single linear layer, applied after a
LayerNorm to the final hidden state $\mathbf{s}^{(2)}_i$ of the second folding pass, outputs nine
numbers per residue:
the position $\mathbf{t}_i$ and two vectors that are made orthonormal to give $R_i$. The three
backbone atoms are then placed by moving a fixed, idealised template $\mathbf{u}_a$ into that
frame, $\mathbf{p}_{i,a} = R_i\,\mathbf{u}_a + \mathbf{t}_i$ for
$a \in \{\mathrm{N}, \mathrm{C}\alpha, \mathrm{C}\}$. The whole head has about 45 thousand
parameters, against the roughly 700 million fine-tuned in the 8-billion-parameter language model.
There is no pair representation, no
distogram and no diffusion: the structure is a linear readout of the language model's hidden
states.

\paragraph{Training objective.}
We follow the standard structure loss. The main term is the frame-aligned point error (FAPE) of
AlphaFold2 \citep{jumper2021highly}, which compares predicted and true atom positions in each
residue's local frame and so does not depend on how the structure is rotated or shifted. We
compute it within chains and between chains, and add two small terms that penalise unrealistic
bond lengths between consecutive residues and clashes between atoms. The exact form and weights
are in \cref{app:loss}.

\subsection{Training}
\label{sec:stages}

\begin{figure}[t]
  \centering
  \includegraphics[width=\linewidth]{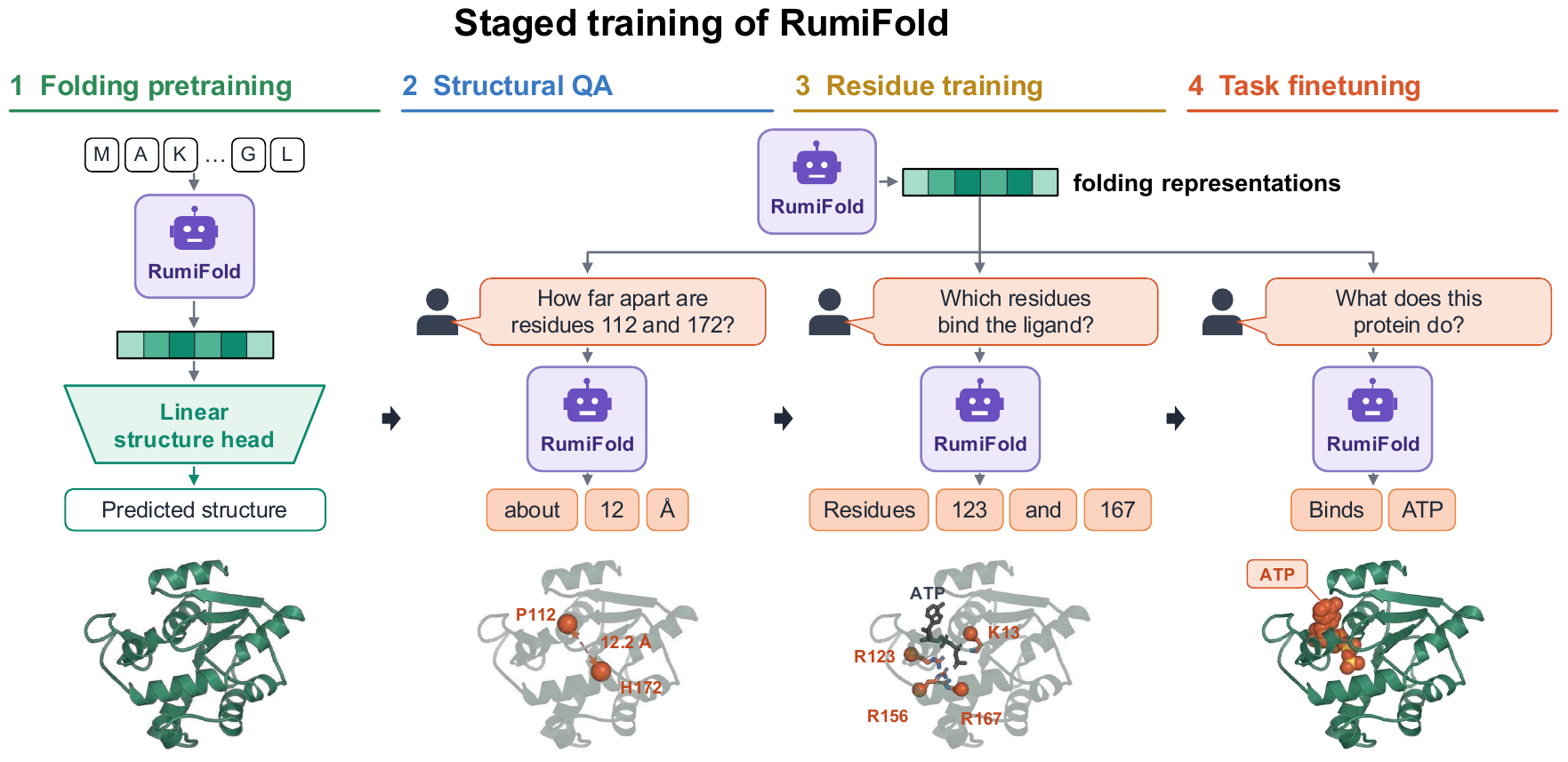}

\caption{\textbf{RumiFold learns to fold, understand, and describe proteins through four stages of training.}
  \emph{Stage~1 (Folding, updates 0--9k):} The model learns to predict protein structure, developing structural representations in its hidden states.
  \emph{Stage~2 (Structural QA, 9k--41k):} The model learns to access these representations to answer questions about its own predicted structure in natural language.
  \emph{Stage~3 (Functional training, 41k--95k):} The model learns to connect its structural representations to biological function.
  \emph{Stage~4 (Task fine-tuning):} The model is adapted to downstream tasks, including function prediction, residue annotation, and protein question answering.
  The stages are cumulative: each adds a task to those before it, so the same language model keeps training on folding through all four.}
  \label{fig:stages}
\end{figure}

\paragraph{Folding data.}
We trained on PDB biological assemblies and AlphaFold Database models. For the PDB set, we used a release-date cutoff of January 1, 2022, and selected structures determined by X-ray crystallography or cryo-EM at a resolution of 3.5\,\AA{} or better, with at least 30 residues per chain. For the AlphaFold Database set, we selected models with a global pLDDT of at least 70. Together, the two sources yielded about 1.6 million assemblies, each containing at most 2{,}048 residues across all chains.

\paragraph{Two models.}
We report two trained models. \emph{\ours{}} is trained through the stages below with language
replay throughout; it is the model that answers questions, and we also evaluate its folding.
\emph{\foldonly{}} is trained only to fold, for 176k updates; we use it as a structure-only reference in the
folding benchmarks (\cref{sec:fold}) and in the head analysis (\cref{sec:interp}).

\paragraph{Stages.}
\ours{} is trained in the four stages of \cref{fig:stages}. The stages are cumulative: each adds a
task while the earlier ones continue, so folding is trained from the first update to the last. The
step that makes folding useful for language is stage~2. There, the answers to
structural questions (the distance between two residues, how buried a residue is, which residues
are in contact) are computed on the model's \emph{own predicted} coordinates, rather than the native
structure. The model is therefore never rewarded for reciting a fact about the protein; it is
rewarded for reporting what its fold says. The transitions are gated on performance and not on a
fixed schedule: structural question answering begins once held-out folding accuracy reaches an
\lddt{} of about 0.7, functional training once it reaches about 0.8 and structural questions are
answered reliably, and task fine-tuning after roughly one pass through the functional training
data. The question types and the perturbed folds used late in stage~2 are described in
\cref{app:perturb}, the answer format of stage~3 in \cref{app:qual}, and the training data and
its held-out sets in \cref{app:data}.

\paragraph{Preserving language.}
Replay prompts are drawn from the Tulu-3 mixture \citep{lambert2024tulu}, with responses regenerated by base
Qwen3-8B (823k conversations). On these turns the model is trained with a KL term toward base
Qwen's next-token distribution plus a small cross-entropy term, half of the time with a protein in
context.

\paragraph{Compute.}
Stages 1 to 3 of \ours{} were trained for about three weeks on 8 A100 GPUs. Each task fine-tune of stage 4 (Gene Ontology, Prot2Text) took about ten days on 4 H100 GPUs. Optimisation details are in
\cref{app:optim}.

\section{Has the LLM Learned to Fold?}
\label{sec:fold}

\begin{figure}[t]
  \centering
  \begin{tikzpicture}
    \begin{axis}[paper, name=fbench, ybar=0.6pt, area legend, bar width=6.6pt, width=0.55\linewidth, height=4.4cm,
        title={FoldBench monomers (334 targets)}, symbolic x coords={TM,lDDT,GDT}, xtick=data,
        xticklabels={TM-score, CA-\lddt{}, GDT-TS}, enlarge x limits=0.24,
        ymin=0.5, ymax=0.98, ytick={0.5,0.6,0.7,0.8,0.9}, minor y tick num=1, yminorgrids,
        minor grid style={gray!10},
        legend to name=foldlegend, legend columns=3]
      \addplot[reference bar] coordinates {(TM,0.871) (lDDT,0.850) (GDT,0.827)};
      \addplot[fill=cGrey, draw=none, vals=3] coordinates {(TM,0.709) (lDDT,0.659) (GDT,0.591)};
      \addplot[fill=cBaseB, draw=none] coordinates {(TM,0.783) (lDDT,0.747) (GDT,0.710)};
      \addplot[fill=cBaseA, draw=none, vals=3] coordinates {(TM,0.832) (lDDT,0.802) (GDT,0.765)};
      \addplot[fill=cOurs, draw=none] coordinates {(TM,0.840) (lDDT,0.781) (GDT,0.756)};
      \addplot[fill=cCtrl, draw=none, vals=3] coordinates {(TM,0.846) (lDDT,0.792) (GDT,0.767)};
      \legend{ESMFold (pair rep.), DPLM-2 3B, ESM3-open, SimpleFold 700M, \ours{}, \foldonly{}}
    \end{axis}
    \begin{axis}[paper, at={(fbench.south east)}, xshift=0.85cm, ybar=0.6pt, area legend, bar width=6.6pt,
        width=0.55\linewidth, height=4.4cm,
        title={CASP16 (155 domains)}, symbolic x coords={TM,lDDT,GDT}, xtick=data,
        xticklabels={TM-score, CA-\lddt{}, GDT-TS}, enlarge x limits=0.24,
        ymin=0.4, ymax=0.88, ytick={0.4,0.5,0.6,0.7,0.8}, minor y tick num=1, yminorgrids,
        minor grid style={gray!10}]
      \addplot[reference bar] coordinates {(TM,0.764) (lDDT,0.767) (GDT,0.710)};
      \addplot[fill=cGrey, draw=none, vals=3] coordinates {(TM,0.547) (lDDT,0.592) (GDT,0.466)};
      \addplot[fill=cBaseB, draw=none] coordinates {(TM,0.651) (lDDT,0.668) (GDT,0.588)};
      \addplot[fill=cBaseA, draw=none, vals=3] coordinates {(TM,0.706) (lDDT,0.730) (GDT,0.641)};
      \addplot[fill=cOurs, draw=none] coordinates {(TM,0.714) (lDDT,0.706) (GDT,0.636)};
      \addplot[fill=cCtrl, draw=none, vals=3] coordinates {(TM,0.720) (lDDT,0.714) (GDT,0.642)};
    \end{axis}
  \end{tikzpicture}\\[-2pt]
  \pgfplotslegendfromname{foldlegend}
  \caption{\textbf{A frozen language model with adapters folds about as well as a dedicated
    single-representation model.} Single-sequence structure prediction on FoldBench monomers and
    CASP16 domains: C$\alpha$ TM-score, C$\alpha$-\lddt{} and GDT-TS. \ours{} is the model that
    also answers questions; \foldonly{} is trained only to fold. DPLM-2, ESM3 and SimpleFold are
    single-representation models. ESMFold (hatched) uses
    pair representations and is shown as a reference.}
  \label{fig:fold}
\end{figure}
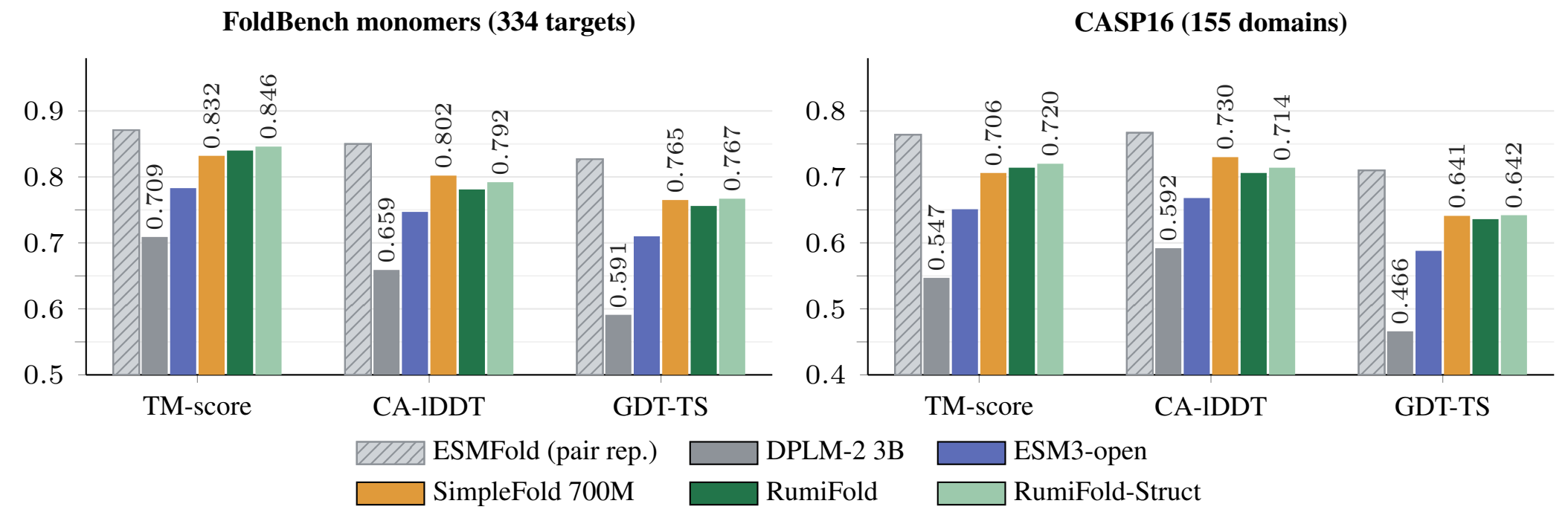

We evaluate \ours{} itself, the model that goes on to answer questions, and compare it with
three single-representation models, ESM3-open
\citep{hayes2025simulating}, DPLM-2 3B \citep{wang2024dplm} and SimpleFold 700M \citep{wang2025simplefold}, and with ESMFold
\citep{lin2023evolutionary}. We use the 334 monomer targets of FoldBench
\citep{xu2025benchmarking} and the 155 evaluation domains of CASP16 \citep{kryshtafovych2025progress}, and score TM-score,
\lddt{} and GDT-TS on C$\alpha$ coordinates.

On FoldBench \ours{} reaches a TM-score of 0.840, above SimpleFold 700M (0.832), ESM3-open
(0.783) and DPLM-2 3B (0.709; \cref{fig:fold}). On CASP16 it reaches 0.714, against 0.706 for
SimpleFold 700M. \foldonly{}, which spends all of its capacity on structure, is a little better
again (0.846 and 0.720). SimpleFold 700M is more accurate locally: its \lddt{} is higher by 0.02
on both sets (0.802 against 0.781, and 0.730 against 0.706), and its GDT-TS by 0.01. So \ours{}
places the chain as well as a dedicated single-representation model of comparable trained size
and is a little less exact in the local detail. The largest SimpleFold, at 3B parameters, has the
same TM-score as \ours{} on FoldBench and is 0.01 behind on CASP16 (\cref{tab:calpha}). We
also evaluate on CAMEO22 and CASP14, which we screened out of training at 30\% sequence identity,
and report these held-out results in \cref{app:fold}; there SimpleFold 700M is ahead of \ours{}. Because \ours{} takes two chains, it also predicts complexes, which the other
single-representation models cannot (\cref{app:complex}).
We also ask whether the language model's text pretraining matters for folding, or whether any
transformer of this size would do, by comparing pretrained, shuffled and randomly initialised
weights (\cref{sec:lm}). We find that language pretraining helps in our setting, where the base
model is frozen and only adapters are trained. Inside the trained model, folding is carried by a
small set of attention heads that track residue contacts, and these overlap with heads that match
repeated tokens in base Qwen (\cref{sec:interp}).

\section{Is Folding Knowledge Useful for Predicting Function?}
\label{sec:function}

We now turn to the question the folding stage was built for. A model that folds has not
necessarily learned anything it can use elsewhere, so we evaluate \ours{} on two function
benchmarks, Gene Ontology prediction and free-text description. On every benchmark a large share of test proteins
has a close relative in the training set, and for those proteins recall is enough. We therefore report results by sequence identity or novelty from the training set.

\subsection{Gene Ontology prediction}
\label{sec:go}

\paragraph{Setup.}
We predict Gene Ontology terms \citep{ashburner2000gene} for molecular function, biological
process and cellular component from sequence, on the benchmark released with GO-GPT v2
\citep{fallahpour2026bioreason}. All of our models are trained on its training set (169{,}810
labelled proteins) and evaluated on its \texttt{proteins} test set with its tier-1 labels and
information-weighted F1 \citep{zhou2019cafa}.
We report two of our models. \emph{\ours{}} writes its GO terms as text, after functional training
and fine-tuning on this benchmark. \emph{\foldonly{} + GO head} is the structure-only model
fine-tuned with a classification head that reads only the language model's hidden states. Both keep a small folding loss on
during function training, and both still fold afterwards: on FoldBench, function fine-tuning
leaves the TM-score of \ours{} unchanged ($+0.001$, $p = 0.65$) and lowers that of the latter by
0.007.

\paragraph{Results.}
\ours{} reaches a mean weighted F1 of 0.683 and \foldonly{} with a GO head 0.657, against 0.666 for
the released GO-GPT v2 checkpoint and 0.655 for BioReason-Pro given that checkpoint's predictions
(\cref{tab:go}); the difference from GO-GPT v2 is significant ($p = 0.0003$, paired permutation
test over test proteins). Against the released
GO-GPT v2, \ours{} is ahead on molecular function (0.685 against 0.665, $p = 0.020$) and cellular component
(0.748 against 0.718, $p = 10^{-4}$) and level on biological process (0.615 against 0.614). It gets there with far less training on
the benchmark: the released GO-GPT v2 checkpoint has made 90 passes over the training set, and
\ours{} makes 16, after its folding and functional training stages. The untrained Qwen3-8B, given the sequence as text,
scores 0.110, so none of this comes from the language model's prior knowledge.

\begin{table}[t]
  \caption{\textbf{Gene Ontology prediction on the GO-GPT v2 benchmark.} Information-weighted F1
    for molecular function (MF), biological process (BP) and cellular component (CC) on the
    \texttt{proteins} test set (3{,}726 proteins), all rows scored with the same script.
    All generated answers, ours and the baselines', use greedy decoding: one answer per protein,
    with no sampling and no selection among candidates. BioReason-Pro reads the predictions of the released
    GO-GPT v2 in its prompt. Bold marks the best model and underline the second best.
    \emph{$<$50\%}: mean over test proteins below 50\% identity to any training protein.
    \emph{Right:} the same score on the 181 test proteins below 30\% identity to any training
    protein, by aspect (BioReason-Pro RL, GO-GPT v2 and \ours{}; the axis starts at 0.3).}
  \label{tab:go}
  \centering
  \begin{minipage}[c]{0.61\linewidth}
  \centering
  \small
  \resizebox{\ifdim\width>\linewidth\linewidth\else\width\fi}{!}{%
  \begin{tabular}{@{}lcccc@{\hspace{10pt}}c@{}}
    \toprule
    Model & MF & BP & CC & Mean & $<$50\% \\
    \midrule
    Qwen3-8B, untrained & 0.111 & 0.058 & 0.161 & 0.110 & 0.110 \\
    BioReason-Pro (SFT) & 0.617 & 0.571 & 0.679 & 0.622 & \underline{0.511} \\
    BioReason-Pro (RL)  & 0.649 & 0.600 & 0.714 & 0.655 & 0.496 \\
    GO-GPT v2 (released) & 0.665 & \underline{0.614} & \underline{0.718} & \underline{0.666} & \underline{0.511} \\
    \midrule
    Qwen + \esmc{} (no fold pretraining)                    & 0.664 & 0.537 & 0.712 & 0.638 & 0.480 \\
    Qwen + ESMFold2 states                        & 0.633 & 0.577 & 0.705 & 0.638 & 0.431 \\
    \foldonly{} + GO head                           & \underline{0.675} & 0.590 & 0.706 & 0.657 & \textbf{0.521} \\
    \ours{}                                         & \textbf{0.685} & \textbf{0.615} & \textbf{0.748} & \textbf{0.683} & 0.508 \\
    \bottomrule
  \end{tabular}}
  \end{minipage}\hfill
  \begin{minipage}[c]{0.36\linewidth}
    \centering
    \begin{tikzpicture}
      \begin{axis}[paper, ybar=0.6pt, area legend, bar width=7pt, width=\linewidth, height=4.3cm,
          symbolic x coords={MF,BP,CC}, xtick=data, enlarge x limits=0.25,
          ymin=0.3, ymax=0.65, ytick={0.3,0.4,0.5,0.6}, ylabel={F1 below 30\% identity},
          legend style={at={(0.03,0.98)}, anchor=north west, font=\scriptsize}, legend columns=1]
        \addplot[fill=cBaseB, draw=none] coordinates {(MF,0.456) (BP,0.326) (CC,0.540)};
        \addplot[fill=cBaseA, draw=none] coordinates {(MF,0.469) (BP,0.345) (CC,0.553)};
        \addplot[fill=cOurs,  draw=none] coordinates {(MF,0.466) (BP,0.355) (CC,0.609)};
        \legend{BioReason-Pro, GO-GPT v2, \ours{}}
      \end{axis}
    \end{tikzpicture}
  \end{minipage}

\end{table}

\paragraph{What the published systems add.}
Two observations put the baselines in context. First, GO-GPT v2's accuracy on biological process
climbs late in its training (0.432, 0.504 and 0.592 at epochs 25, 50 and 75) and its by-identity
profile shows those gains arriving in the highest-identity bin, the signature of accumulating
recall. Second, BioReason-Pro's accuracy tracks the GO-GPT predictions it is given in its prompt
and does not exceed them: given the predictions of the released GO-GPT v2 checkpoint it scores
0.655 (RL) and 0.622 (SFT), below those predictions alone (0.662; they come from the released checkpoint run on
BioReason-Pro's copies of the test sequences, and it scores 0.666 on the benchmark's own, \cref{tab:go}). This agrees with the finding of
\citet{fang2026when} that the model's answers follow its text inputs.
\ours{} also performs strongly when fine-tuned on the original GO-GPT v1 training set: scored on that benchmark's test set, it reaches 0.673, against
0.664 for GO-GPT v1 and 0.662 for BioReason-Pro (\cref{app:brp}).

\paragraph{Is it folding, or just \esmc{}?}
\label{sec:controls}
\ours{} reads \esmc{} embeddings, whereas GO-GPT v1 and v2 read ESM-2 (3B and 650M), so its gains
could come from the newer encoder and not from learning to fold. Three controls address this (\cref{tab:go}).
\emph{The same model without folding.} \emph{Qwen + \esmc{}} is base Qwen3-8B with the same
bridge, LoRA rank and training data, whose adapters were never trained to fold. It scores 0.638
against 0.683 for \ours{} ($p = 10^{-4}$), and 0.480 against 0.508 below 50\% identity ($p = 0.0065$).
\emph{A newer encoder is not enough.} We retrain GO-GPT v2 with its exact recipe, changing only
the encoder from ESM-2 to \esmc{}-6B. At the released checkpoint's epoch (90) it scores 0.651, below the released model (0.666).
\emph{A folding model's states in place of folding.} We give base Qwen3-8B, with the same LoRA
rank, the trunk states of frozen ESMFold2 and train it on the same data. It reaches 0.638 against
0.683 for \ours{} ($p = 10^{-4}$), and the gap is larger below 50\% identity (0.431 against 0.508, $p = 10^{-4}$). Reading a dedicated folding model's states is therefore not a
substitute for the language model learning to fold.

\subsection{Free-text function description}
\label{sec:p2t}

\begin{figure}[t]
  \centering
  \begin{minipage}[b]{0.50\linewidth}
    \centering
    \small
    \setlength{\tabcolsep}{4pt}
    \begin{tabular}{@{}lccc@{}}
      \toprule
      Model & BLEU-4 & R-L & GPT-4o \\
      \midrule
      \multicolumn{4}{@{}l}{\emph{All 4{,}203 test proteins}} \\
      SEQ2LLaMA & 25.8 & 35.2 & 35.2 \\
      Prot2Text-V2 & 39.0 & \underline{55.7} & \underline{59.5} \\
      \ours{}      & \underline{41.0} & 54.5 & 58.3 \\
      \ours{} (L76) & \textbf{42.5} & \textbf{55.8} & \textbf{60.2} \\
      \midrule
      \multicolumn{4}{@{}l}{\emph{2{,}063 novel proteins}} \\
      SEQ2LLaMA & 9.9 & 16.0 & 15.6 \\
      Prot2Text-V2 & 15.9 & 26.4 & 31.9 \\
      \ours{}      & \underline{19.3} & \underline{27.0} & \underline{32.4} \\
      \ours{} (L76) & \textbf{21.1} & \textbf{28.0} & \textbf{34.4} \\
      \bottomrule
    \end{tabular}
    \vspace{8pt}
    \subcaption{BLEU-4, ROUGE-L and GPT-4o judge score (0--100). Bold: best; underline: second. \ours{} (L76) reads layer 76 of \esmc{} in place of the final layer.}
    \label{fig:p2t-table}
  \end{minipage}\hfill
  \begin{minipage}[b]{0.47\linewidth}
    \centering
    \begin{tikzpicture}
      \begin{axis}[paper, ybar, area legend, bar width=7pt, width=\linewidth, height=5cm, xmin=0.3, xmax=7.4,
          xtick={1,2,3,4.7,5.7,6.7}, xticklabels={Prot2Text-V2,\ours{},\ours{} (L76),Prot2Text-V2,\ours{},\ours{} (L76)},
          x tick label style={font=\tiny, rotate=30, anchor=north east, inner sep=1pt},
          ymin=0, ymax=58, ytick={0,10,20,30,40,50}, ylabel={BLEU-4}, vals=1,
          legend style={at={(0.5,-0.42)}, anchor=north}, legend columns=2]
        \node[font=\scriptsize\itshape] at (axis cs:2,55) {all 4{,}203};
        \node[font=\scriptsize\itshape] at (axis cs:5.7,55) {2{,}063 novel};
        \addlegendimage{fill=rumiGrey, draw=none}\addlegendentry{trained model}
        \addlegendimage{reference bar}\addlegendentry{lookup in its input embedding}
        \addplot[fill=cBaseA, draw=none, bar shift=-4pt, forget plot] coordinates {(1,39.0) (4.7,15.9)};
        \addplot[fill=cOurs, draw=none, bar shift=-4pt, forget plot] coordinates {(2,41.0) (5.7,19.3)};
        \addplot[fill=cCtrl, draw=none, bar shift=-4pt, forget plot] coordinates {(3,42.5) (6.7,21.1)};
        \addplot[reference bar, bar shift=4pt, forget plot] coordinates {(1,42.9) (2,33.5) (3,45.9) (4.7,24.3) (5.7,20.0) (6.7,26.8)};
      \end{axis}
    \end{tikzpicture}
    \subcaption{BLEU-4 of each trained model (solid) beside a lookup in the embedding it reads (hatched): copying the description of the nearest training protein, with no training. On all test proteins, \ours{} is the only model that beats its lookup (41.0 against 33.5).}
    \label{fig:p2t-lookup}
  \end{minipage}
  \caption{\textbf{\ours{} describes novel proteins better than Prot2Text-V2, and reading a
    better \esmc{} layer extends the lead.} All models are run on the same test proteins with the same prompt (no protein
    name or taxonomy), greedy decoding and the same scorers. \emph{Novel} proteins are those
    whose reference description does not occur in the training set. In (b), the lookup
    uses the mean over residues and cosine distance; Prot2Text-V2 reads the final layer of ESM-2
    3B, \ours{} the final layer of \esmc{} and \ours{} (L76) its layer 76.}
  \label{fig:p2t}
\end{figure}
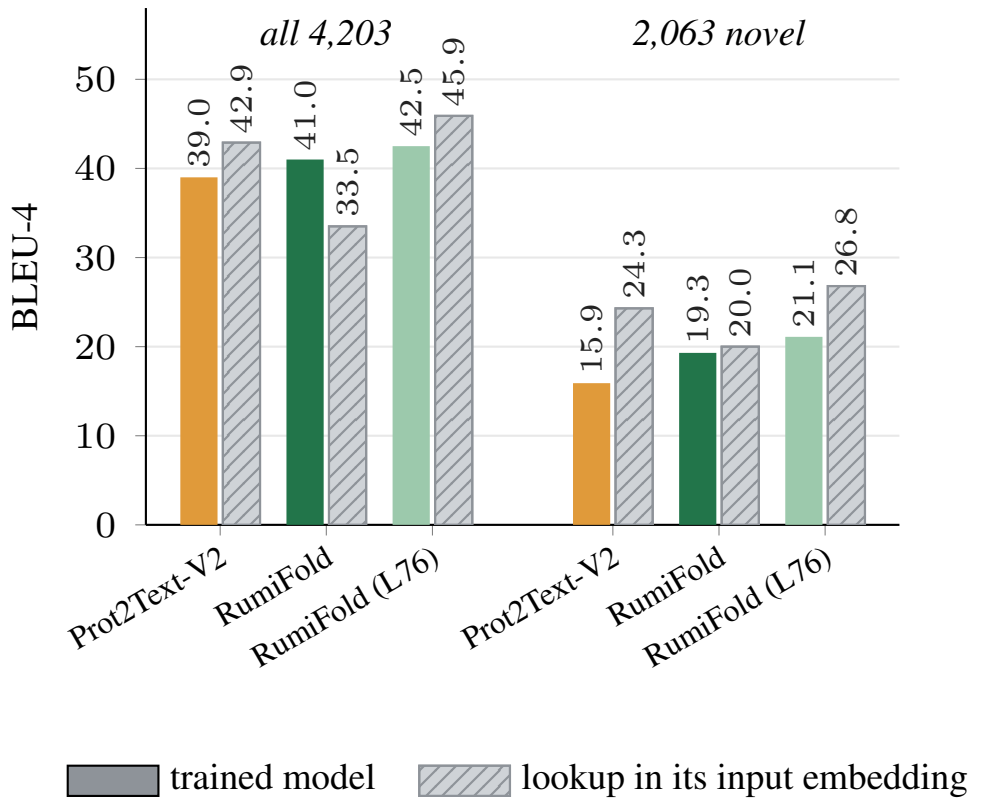

Prot2Text \citep{fei2025prot2text} asks for a free-text description of a protein's function. We
fine-tune \ours{} on its training set and evaluate on its 4{,}203 test proteins, and separately
on the 2{,}063 \emph{novel} proteins whose reference description does not occur in the training
set. We compare with Prot2Text-V2, the strongest published system, by
running its released weights on the same proteins with the same bare prompt, greedy decoding
and the same scorers.

We report three measures. BLEU-4 is the precision of the generated one- to four-word phrases
against the reference description, with a penalty for answers shorter than it. ROUGE-L
\citep{lin2004rouge} is an F-score on the longest word sequence the two share in order. Both
count shared words, so we also report the judge score of the Prot2Text-V2 paper: GPT-4o reads
the reference and the generated description and rates from 0 to 10 how well the prediction
captures the function, its specifics and its completeness; we report the mean on a 0--100 scale.

\paragraph{Results.}
\ours{} describes novel proteins better than Prot2Text-V2 (\cref{fig:p2t-table}). On the
2{,}063 proteins whose descriptions cannot be recalled from training it reaches BLEU-4 19.3
against 15.9, and is level or slightly ahead on ROUGE-L (27.0 against 26.4) and on the judge
score (32.4 against 31.9). Over all test proteins the two are close: \ours{} is ahead on BLEU-4
(41.0 against 39.0) and behind on ROUGE-L (54.5 against 55.7) and the judge score (58.3 against
59.5). It gets there with less training on the benchmark: Prot2Text-V2 is trained for 12 epochs
of contrastive alignment followed by 24 epochs of instruction tuning.

\paragraph{The input layer matters.}
The two models read different encoders, and the encoders differ in how much function they carry
before any training. We measure this with a lookup: each test protein is given the description
of its nearest training protein in the embedding the model reads (\cref{fig:p2t-lookup}). The
final layer of \esmc{}, which \ours{} reads, is a poor description of function on its own: the
lookup scores 33.5 BLEU-4, against 42.9 in the ESM-2 3B features that Prot2Text-V2 reads.
\ours{} is the only model that outperforms the lookup in its own input embedding: over all test
proteins it reaches 41.0 against 33.5, while Prot2Text-V2 generates below its input's lookup
(39.0 against 42.9).

\paragraph{Reading a better layer.}
Earlier layers of \esmc{} carry more: the same lookup in layer 76 scores 45.9, above ESM-2 3B.
We therefore fine-tune \ours{} in the same way on layer 76 in place of the final layer, a choice
made on validation data, and average the weights of its late checkpoints. \ours{} (L76) is then
ahead of Prot2Text-V2 on all three measures over all test proteins (BLEU-4 42.5 against 39.0,
ROUGE-L 55.8 against 55.7, judge score 60.2 against 59.5), though the last two are within noise
(judge difference $+0.7$, 95\% interval $[-0.1, +1.5]$). On novel proteins the lead is clear:
BLEU-4 21.1 against 15.9, ROUGE-L 28.0 against 26.4, and judge score 34.4 against 31.9
(paired difference $+2.5$, 95\% interval $[+1.4, +3.6]$).

An earlier experiment on another free-text description benchmark (Mol-Instructions) is in \cref{app:molinst}, where RumiFold is also state of the art among trained protein models.

\section{What Does Learning to Fold Enable Beyond Structure Prediction?}
\label{sec:mechanism}

What does a language model gain from learning to fold, beyond the ability to predict structure? We investigate this in three steps. First, we examine how folding changes the model's internal representations of protein structure and function. Second, we test whether the language model uses these folding representations as an internal workspace when answering questions. Finally, we ask whether this workspace reflects biologically meaningful mechanisms, connecting the model's predictions to experimentally characterized functional residues.

\subsection{Folding enriches the representation}
\label{sec:enrich}

\begin{figure}[t]
  \centering
  \includegraphics[width=\linewidth]{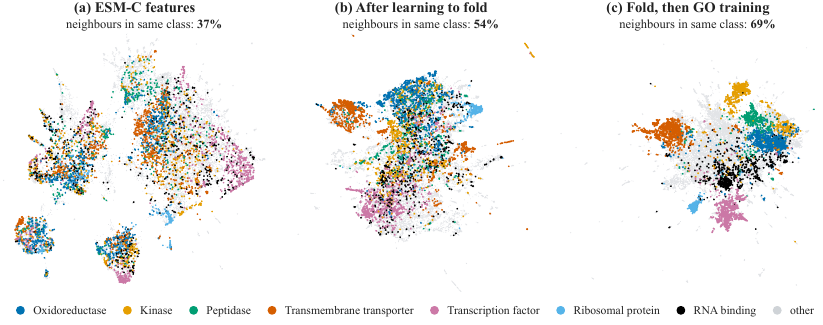}
  \caption{\textbf{Learning to fold separates proteins by function, and function training then
    sharpens it.} UMAP of residue-averaged representations of the
    same 23{,}084 proteins:
    \textbf{(a)}~the final \esmc{} layer the models read;
    \textbf{(b)}~the states of \foldonly{};
    \textbf{(c)}~the same model after Gene Ontology fine-tuning.
    Colours mark seven molecular-function classes by true annotation, identical in every panel;
    proteins in none or in several are grey. The percentage is the share of a test protein's ten
    nearest training proteins in the full space (cosine distance) that carry its class, over 660
    test proteins; chance is 4\%.}
  \label{fig:umapgrid}
\end{figure}

\paragraph{Folding organises proteins by function before any fine-tuning on function data.}

\Cref{fig:umapgrid} draws the same proteins in three representation spaces. In the \esmc{}
features the model is fed, functional classes overlap: a protein's nearest neighbours share its
molecular-function class 37\% of the time. After training only to fold, with no functional label
ever seen, they do so 54\% of the time, and transporters, transcription factors and ribosomal
proteins have separated. Function training on top of folding takes this to 69\%. Lookup of Gene Ontology terms by nearest neighbour tells the same story
(0.435 in \esmc{}, 0.455 after folding, 0.523 after folding and function training; \cref{sec:refine}).

\paragraph{Residue-level probes.}
The same holds residue by residue. We fit one probe on frozen features of \foldonly{} and of
\esmc{}-6B across 17 VenusX residue tasks \citep{tan2025venusx}, with paired bootstrap intervals
over test proteins (\cref{tab:venusx} in \cref{app:venusx}). The fold-only states are ahead on 8 tasks, including
every domain task and the active-site and conserved-site tasks on the MF50 and MP50 splits, and
behind on 2, the cross-family active-site and conserved-site tasks. The remaining 7 are not
separated. The states of the control fine-tuned on function without folding are below the
\esmc{} features on all 17 tasks, and below the fold-only states on 15.

We also show that this enrichment is computed during the folding passes and is not passed through from the
encoder (\cref{app:during}).

\subsection{RumiFold Uses Its Internal Folding Workspace}
\label{sec:workspace}

\begin{figure}[t]
  \centering
  \includegraphics[width=\linewidth]{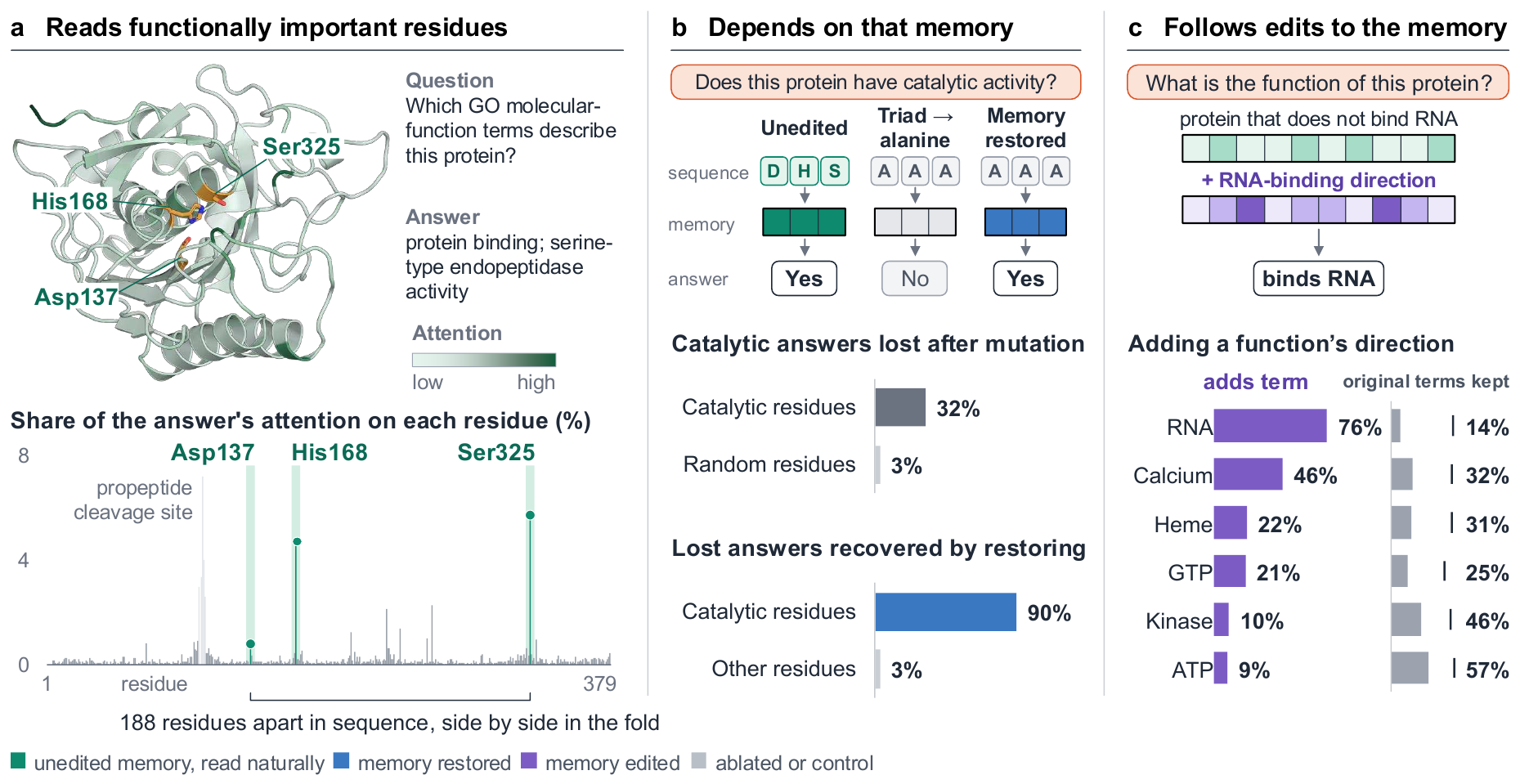}
  \caption{\textbf{\ours{} uses its internal folding workspace.}
    \textbf{(a)}~When \ours{} names the function of subtilisin, its attention falls on the
    catalytic triad (Asp137, His168, Ser325), three residues far apart in the sequence and side
    by side in the fold.
    \textbf{(b)}~Mutating an enzyme's catalytic residues to alanine removes the answer
    ``catalytic'' for 32\% of 1{,}448 held-out enzymes, against 3\% for random residues.
    Restoring the memory at those residues alone recovers 90\% of the lost answers, against 3\%
    for restoring it at other residues.
    \textbf{(c)}~Adding a function's direction to the memory makes the model write that
    function for proteins that lack it (about 2{,}600 per function), at the cost of some of the
    terms it wrote before; ticks show a random direction of the same size.}
  \label{fig:workspace}
\end{figure}

We call the \emph{memory} of a protein the keys and values that the second folding pass computes
for each of its residues at every layer (\cref{eq:question}): one entry per residue and layer,
written while the model folds. By design, the language pass attends to this memory and to nothing else about the protein. In the previous section, we showed that folding organises that memory around structure and function (\cref{sec:enrich}). Here, we ask \emph{how} the model uses this memory. Which residues receive the most attention when
the model names a function? Does restoring the memory at those residues
recover an answer changed by mutation? Can editing a direction in the
value vectors change which functions the model reports? (\cref{fig:workspace})

\paragraph{The answer reads functionally important residues.}
When \ours{} answers a question about a protein's function, its attention is concentrated on
the residues that carry out that function. In subtilisin, the catalytic triad lies far apart in
the sequence and meets only in the fold, and the answer finds it: of 379 residues, Ser325 and
His168 receive the second and third largest shares of attention and Asp137 the seventeenth, and
31\% of all attention falls within 10\,\AA{} of the triad, on 17\% of the residues
(\cref{fig:workspace}a). The single most-attended residue is also functional: it is the last
residue of the propeptide, where the enzyme cleaves itself. The pattern is general. Across 369
held-out enzymes with mutagenesis data, residues whose mutation abolishes activity sit at the
89th percentile of attention within their protein, those whose mutation reduces activity at the
81st, and those whose mutation has no reported effect at the 70th.

\paragraph{The answer depends on that memory.}
The model attends strongly to active sites and catalytic residues, but does its answer actually
depend on them? To test this, we mutate the
annotated catalytic residues of an enzyme, with two neighbours on each side, to alanine, and let
the model encode and fold the mutant from scratch. Across 1{,}448 held-out enzymes that are
called catalytic when unedited, 32\% lose that answer; mutating the same number of other
residues changes 3\% (\cref{fig:workspace}b). Measured on the probability itself, the mutation
lowers $\log P(\text{Yes})$ by 0.62 on average, against 0.03 for the control. The question is
deliberately coarse, and a mutant enzyme still has the fold and the family of an enzyme, so most
answers should survive. However, when we keep the mutated sequence
and restore the original memory at the catalytic residues only, the answer returns: this swap
recovers 90\% of the answers the mutation removed (\cref{tab:e2e}), so that 96\% say yes again, and
$\log P(\text{Yes})$ is within 0.09 of the unedited value. Restoring the same
number of entries elsewhere recovers 3\%, leaving 68\% saying yes, no better than the mutant. A few entries of
the memory, at the residues that do the chemistry, are enough to carry the answer. \Cref{app:e2e} gives every condition.

\paragraph{The answer follows edits to the memory.}
If the answer is read from the memory, writing to the memory should change it. For each of five
molecular functions we take one direction in the memory's value space, the mean of proteins with
the function minus the mean of proteins without it, and add it to the memory of every eligible
held-out protein that lacks the function, about 2{,}600 per function. Asked what the protein
does, the model then writes the added function (\cref{fig:workspace}c): for 76\% of proteins
given the RNA-binding direction and 46\% for calcium binding, down to 9\% for ATP binding, the weakest of the five. A random direction of
the same size never adds the term. The edit changes the answer, but it is not
precise: it also disturbs other answers. The proteins keep only 14 to 57\% of the terms they wrote before the edit, against
84 to 89\% under a random direction.

Together these show that the folding memory is a workspace the language model reasons from. The
answer reads the residues that matter, fails when their memory is wrong, recovers when it is put
back, and changes when it is rewritten. \Cref{app:e2e,app:chain} give every condition and
control, including the attention heads that do the reading.

\subsection{The workspace aligns with biological mechanism}
\label{sec:discovery}

\begin{figure}[t]
  \centering
  \includegraphics[width=\linewidth]{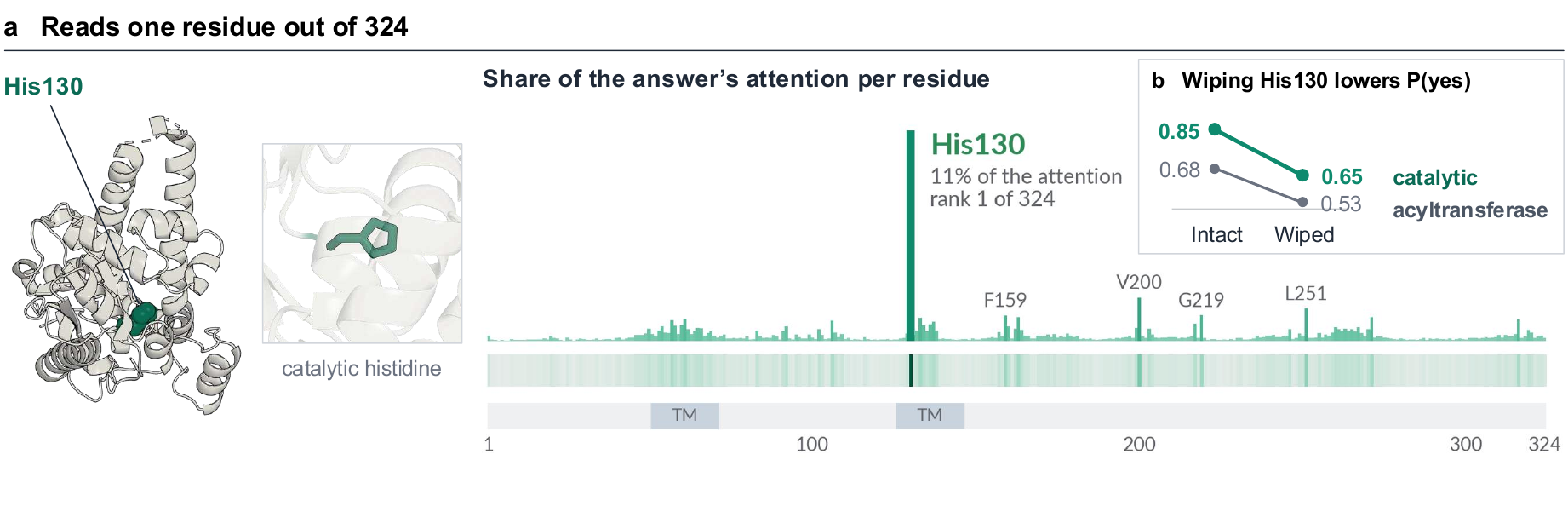}
  \caption{\textbf{\ours{} reads the active site of TMEM68, a protein held out of training.}
    \textbf{(a)}~\emph{Reads one residue out of 324.} When \ours{} names the function of human
    TMEM68, 11\% of the answer's attention is on His130, more than on any other of the 324
    residues. Left: His130, the catalytic histidine, in the predicted structure (AlphaFold
    Database), with a close-up. Right: share of attention along the sequence; TM marks the
    transmembrane segments.
    \textbf{(b)}~\emph{Wiping His130 lowers the answer.} Removing His130 from the folding memory
    lowers the probability of answering yes, from 0.85 to 0.65 for catalytic activity and from
    0.68 to 0.53 for acyltransferase activity.
    TMEM68 and its homologues at 30\% sequence identity or more were excluded from residue and
    function training, and the experiments that identified His130 were published in 2023
    \citep{wang2023tmem68,mclelland2023identification}.}
  \label{fig:tmem68}
\end{figure}

If the answer rests on the residues that carry a function, then reading the model is a way to
ask where a function lives in a protein. We follow one protein through the three steps of
\cref{sec:workspace}, chosen because it and its homologues were held out of training and its
catalytic residue was identified by experiment only in 2023.

\paragraph{A protein whose function was found later.}
Human TMEM68 was known to belong to the glycerophospholipid acyltransferase family, but its
enzymatic function was undetermined until 2023, when it was shown to synthesise triacylglycerol
and to lose that activity when a single histidine, His130, is mutated
\citep{wang2023tmem68,mclelland2023identification}. TMEM68
belongs to the CAFA-style temporal holdout of the GO benchmark: it had no experimental function
annotation before November 2022 and gained one in 2023. It and all homologues at $\geq$30\%
identity were also excluded from our residue and function training. Annotation snapshots for
other training proteins are from 2026, so this is a protein-level holdout rather than a global
knowledge cutoff. The Gene Ontology fine-tuning set does contain its bovine orthologue (93\%
identical), which carries only generic membrane terms and no function or site label.

\paragraph{The model reads His130, and its answer depends on it.}
Asked in plain language whether TMEM68 has catalytic activity, \ours{} answers yes with
probability 0.85, and asked to name its function it writes an acyltransferase activity. His130
receives 11\% of the answer's attention, more than any other of the protein's 324 residues and
about five times the share of the next (\cref{fig:tmem68}a); the same residue is first in the
mouse orthologue. Wiping
that one residue from the folding memory lowers the probability to 0.65, and lowers the more
specific answer, acyltransferase activity, from 0.68 to 0.53 (\cref{fig:tmem68}b). Wiping any
one of 24 control residues, each at least 12\,\AA{} away and equally buried, leaves both answers
unchanged to within the resolution of the readout.

\paragraph{The residue is the one the experiment found.}
The model was given no site annotation for this protein, and it settles on one residue out of the
14 histidines of TMEM68, seven of which lie in its acyltransferase domain. That residue is the one
whose mutation abolishes activity in the laboratory \citep{wang2023tmem68,mclelland2023identification}. His130 is also the residue one would expect from comparing TMEM68 with related enzymes, so the
case is not a discovery. What it shows is that the model's internal workspace, given no site label
for this protein in training, still aligns with the residue that does the chemistry.

This is one selected protein, and we do not claim that the model's attention finds active sites
in general. Across 369 held-out enzymes, wiping the annotated active-site residues moves the
same answer much less than wiping the residues the model attends to most, and attention separates
annotated sites from other residues only about as well as burial does. What the case shows is
that the workspace can be interrogated residue by residue, and that an interrogation can land on
real biochemistry. \Cref{app:cases} gives further cases, among them ones where the readout misses and an
untested hypothesis.

\paragraph{What learning to fold enables.}
Learning to fold gives the language model three things beyond a structure. Folding enriches
its representation: proteins and residues become organised by function before any function
training (\cref{sec:enrich}). The model uses the workspace that folding leaves behind: its
answers read the residues that matter, depend on what is stored there, and follow edits to it
(\cref{sec:workspace}). And that workspace aligns with biological mechanism: in a protein held
out of training, the residue the answer rests on is the one an experiment showed to be
catalytic (\cref{sec:discovery}). \Cref{app:e2e} gives every control for the mutation
experiments and \cref{app:chain} traces single functions through the memory.

\section{Limitations}
\label{sec:limits}

\begin{figure}[t]
  \centering
  \begin{tikzpicture}
    \begin{axis}[paper, name=ret, ybar=0.7pt, area legend, bar width=5.5pt, width=0.58\linewidth, height=4.5cm,
        title={(a) Accuracy change vs.\ parent LLM (reasoning mode)},
        symbolic x coords={ARC-C,TruthfulQA,MMLU,GSM8K}, xtick={ARC-C,TruthfulQA,MMLU,GSM8K}, enlarge x limits=0.17,
        ymin=-40, ymax=0, ytick={-40,-30,-20,-10,0}, ylabel={percentage points},
        nodes near coords,
        every node near coord/.append style={font=\tiny, rotate=90, anchor=east, inner sep=1.5pt,
          /pgf/number format/.cd, fixed, fixed zerofill, precision=1},
        legend style={at={(0.5,-0.2)}, anchor=north}, legend columns=2]
      \addplot[reference bar] coordinates {(ARC-C,-4.7) (TruthfulQA,-3.8) (MMLU,-7.4) (GSM8K,-8.7)};
      \addplot[fill=cOurs,  draw=none] coordinates {(MMLU,-23.8) (ARC-C,-14.5) (TruthfulQA,-21.0) (GSM8K,-14.0)};
      \addplot[fill=cBaseB, draw=none] coordinates {(MMLU,-23.2) (ARC-C,-19.0) (TruthfulQA,-32.5) (GSM8K,-27.5)};
      \addplot[fill=cBaseA, draw=none] coordinates {(MMLU,-21.8) (ARC-C,-21.0) (TruthfulQA,-29.5) (GSM8K,-29.5)};
      \legend{{replay only, no proteins}, \ours{}, BioReason-Pro (SFT), BioReason-Pro (RL)}
    \end{axis}
    \begin{axis}[paper, at={(ret.south east)}, xshift=1.3cm, ybar=0.7pt, area legend, bar width=8pt,
        width=0.40\linewidth, height=4.5cm, title={(b) Wikipedia NLL ($\downarrow$)},
        symbolic x coords={A,B,C}, xtick=data,
        xticklabels={\ours{}, {BioReason-\\Pro (SFT)}, {BioReason-\\Pro (RL)}}, x tick label style={font=\tiny, align=center},
        enlarge x limits=0.28, ymin=2.5, ymax=4.4, ytick={2.5,3,3.5,4}, vals=2,
        legend style={at={(0.5,-0.3)}, anchor=north}, legend columns=2]
      \addplot[fill=cGreyL, draw=none] coordinates {(A,2.94) (B,3.10) (C,3.10)};
      \addplot[fill=cGrey,  draw=none] coordinates {(A,3.65) (B,3.97) (C,3.94)};
      \legend{parent LLM, protein model}
    \end{axis}
  \end{tikzpicture}
  \caption{\textbf{What protein training costs in general language ability.}
    \textbf{(a)}~Change in accuracy relative to each model's own parent LLM, with step-by-step
    reasoning, on 200-item subsets (intervals of about $\pm 6$ points).
    \textbf{(b)}~Per-token negative log-likelihood (nats) on 1{,}578 WikiText passages, about
    259k tokens, each model reading the text inside its own chat template.
    \ours{} is the 95k-step checkpoint in both panels. Hatched bars in (a): the base model
    trained on the replay objective alone, with no protein data, for about 44k steps
    (few-shot direct answering on the full test sets, so compare with the other bars for scale
    only).}
  \label{fig:retention}
\end{figure}
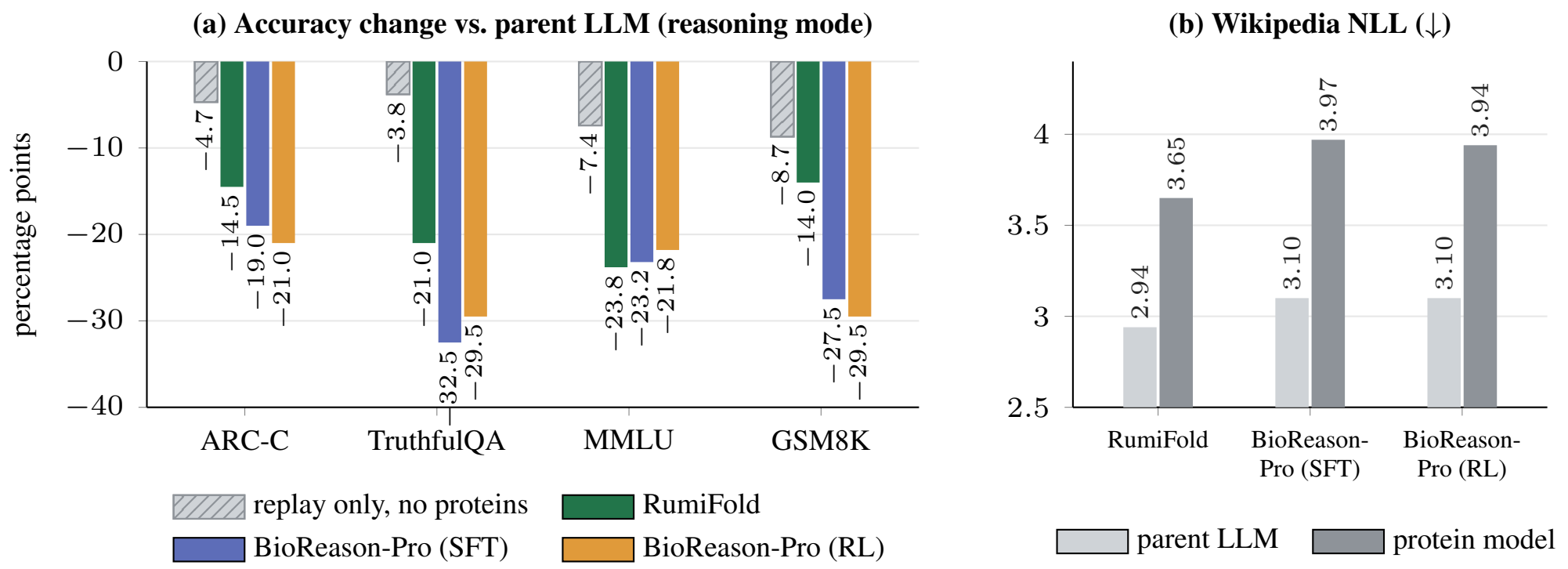

\paragraph{One layer in.}
\ours{} is fed only the final layer of \esmc{}. We chose it at the outset because it is the
standard input, the one that GO-GPT and Prot2Text-V2 take from their encoders. A mean or a learned softmax mixture over the encoder's
layers would be a better input: the final layer is the weakest of those we tested for retrieval
(\cref{sec:function}), and intermediate layers carry more functional information
(\cref{app:venusx}). We did not have the storage or the compute to keep every layer's features
for the folding corpus, and leave this to future work.

\paragraph{What it costs.}
Teaching a language model to compute over proteins in its own weights is not free. Despite replay
of ordinary conversation throughout training, \ours{} loses 14 to 21 points relative to the base
model on MMLU \citep{hendrycks2020measuring}, ARC-Challenge \citep{clark2018think}, TruthfulQA
\citep{lin2021truthfulqa} and GSM8K \citep{cobbe2021training}, a loss comparable to that of
BioReason-Pro against its own parent (\cref{fig:retention}). Part of the cost comes from the
replay itself: the same adapters trained on the replay objective alone, with no protein data,
lose 4 to 9 points on the same four benchmarks. \Cref{tab:retention} gives every number.

\paragraph{Is the hidden state necessary?}
We have argued for processing a protein in the model's hidden states. A different route is opening
up: recent work shows that a model's weights can be edited through text alone \citep{takehi2026weightspace},
which raises the possibility that some of what we obtain here through hidden-state
representations could be reached by working in text. We have not compared the two, and we see
them as complementary directions.

\paragraph{A fixed number of passes.}
\ours{} always folds in two passes. It would make sense to let the number of passes vary, and to
loop: fold, reason in text, then fold again. Fixing the number at two also leaves the first pass
of limited use.
\section{Discussion}
\label{sec:conclusion}

We set out to ask whether a general-purpose language model can learn to fold proteins, and what it gains if it does. RumiFold shows that it can: a frozen Qwen3-8B with low-rank adapters and a linear output head folds proteins about as well as dedicated single-representation models. The folding is useful beyond structure: after folding pretraining, RumiFold reaches state-of-the-art results on Gene Ontology protein function prediction and free-text function description, ahead of matched models that never learned to fold. And the model uses what folding builds: its answers read from the per-residue memory that the folding passes leave behind, depend on it, and change when it is edited.

RumiFold suggests a broader approach to building scientific language models: using foundational scientific tasks as pretraining to develop internal representations that support downstream reasoning. Folding is one computation among many. Evolutionary inference, design and binding could be learned in the same way and would leave their own workspaces. We also imagine models in which language and biomolecular computation alternate. Such a model could run experiments inside its own reasoning: propose a mutation, fold it, read what changed, and decide what to try next. Its chain of thought would contain structures as well as sentences. Harder questions would earn more of these internal experiments, so that thinking longer means computing more, not only writing more.

We see RumiFold as a first step toward \emph{native biomolecular intelligence}: language models in which biomolecular computation and reasoning in language are not separate systems joined by an interface, but capabilities of one model that work together.

\clearpage
\begin{ack}
  This material is based upon work supported by the National Science Foundation Graduate
Research Fellowship Program under Grant No.\ DGE-2439018. Any opinions,
findings, and conclusions or recommendations expressed in this material are those of the
author(s) and do not necessarily reflect the views of the National Science Foundation. We thank Goodfire for supporting this work through a Goodfire Research Grant and for providing access to Silico, their interpretability research agent. This work was also supported by the Alfred P. Sloan Foundation. We thank members of the Bau and Jin labs for their feedback and Sheng-Yu Wang for his thoughtful suggestions on the figures.

\end{ack}

\bibliographystyle{plainnat}
\bibliography{references}

\appendix

\newcommand{\qhead}[1]{\textbf{\small #1}\par\vspace{1pt}}
\newcommand{\qbox}[1]{\par\noindent\fcolorbox{gray!45}{gray!5}{\parbox{\dimexpr\linewidth-2\fboxsep-2\fboxrule}{\footnotesize #1}}\par\vspace{4pt}}
\newcommand{\qlab}[1]{\textcolor{cOurs}{\textbf{#1}}}
\newcommand{\qref}[1]{\textcolor{gray}{\textbf{#1}}}
\newcommand{\gt}{\,$\triangleright$\,}

\FloatBarrier
\section{Architecture and Training Details}
\label{app:archdetails}

\paragraph{Bridge.} \Cref{eq:input} omits one term. In full, the input to the trunk is
\begin{equation}
  \mathbf{h}^0_i = \lambda\,\Big(\mathrm{LayerNorm}\big(W\,\mathrm{LayerNorm}(\mathbf{e}_i) + \mathbf{b}\big)
     + \mathbb{1}[\,i \text{ starts a chain}\,]\;\mathbf{c}\Big),
  \label{eq:bridge}
\end{equation}
where $\mathbf{c}$ is a learned vector that marks the first residue of every chain.

\paragraph{Recycling.} The first folding pass is run without gradient: in \cref{eq:recycle},
$\mathbf{s}^{(1)}_i$ is treated as a constant, so the folding loss trains the trunk through the
second pass only. The model recycles once.

\paragraph{Positions.} Rotary position indices follow the residue numbering within a chain and
jump by 200 between chains; the same indices are used in both folding passes and in the question
pass. Text positions continue directly after the last residue.

\paragraph{Context layout in the question pass.} The opening token of Qwen's chat template
occupies position 0, before the protein, and the rest of the template (a single user turn with
thinking disabled, no system prompt) follows the protein. Text tokens attend to this opening
token as well as to the residues.

\paragraph{Cached keys and values.} The protein entries of \cref{eq:question} are produced by the
trunk's own layer-$\ell$ modules (pre-attention normalisation, key and value projections with
their LoRA adapters, and rotary encoding) applied to the input of block $\ell$ in the second
folding pass, for all 36 blocks. Nothing is added to them. Gradients from the text loss flow
through the cache into the second folding pass.

\paragraph{Residue pointers.} When a question names a residue by its address (for example
``A/129''), that residue's vector is inserted as an extra input embedding directly after the
address. The vector is \cref{eq:recycle} applied to the final hidden state of the second pass in
place of the first. Function questions contain no addresses and are unaffected.

\subsection{Attention masks}
\label{app:masks}

\begin{figure}[h]
  \centering
  \begin{tikzpicture}[sch, x=3.3mm, y=3.3mm]
  \tikzset{cell/.style={draw=white, line width=0.6pt}}
  \foreach \r in {0,...,5} \foreach \c in {0,...,5}
    \fill[cell, fill=rumiGreen] (\c,-\r) rectangle ++(1,-1);
  \foreach \i/\aa in {0/K,1/I,2/W,3/L,4/B,5/B} {
    \node[font=\tiny\ttfamily, anchor=south] at (\i+0.5,0.1) {\aa};
    \node[font=\tiny\ttfamily, anchor=east] at (-0.15,-\i-0.5) {\aa};
  }
  \node[head, anchor=south] at (3,4.3) {(a) Folding passes ($\times 2$)};
  \node[lab, anchor=north, text width=30mm] at (3,-6.4) {every residue attends to every residue; pass 1 is recycled into pass 2};
  \draw[foldflow] (-1.1,-4.3) .. controls +(-2.4,0) and +(-2.4,0) .. node[lab, left, text=rumiGreen] {recycle} (-1.1,-1.7);
  \draw[foldflow] (7.2,-3) -- node[lab, above, text=rumiGreen] {keys and values\\[-1pt]of pass 2, every layer} (12.4,-3);
  \begin{scope}[xshift=52mm]
    \foreach \r in {0,...,4} \foreach \c in {0,...,5}
      \fill[cell, fill=rumiMint] (\c,-\r-0.5) rectangle ++(1,-1);
    \foreach \r in {0,...,4} \foreach \c in {0,...,4} {
      \ifnum\c>\r \fill[cell, fill=rumiGreyL!45] (\c+6.6,-\r-0.5) rectangle ++(1,-1);
      \else \fill[cell, fill=rumiBlue] (\c+6.6,-\r-0.5) rectangle ++(1,-1); \fi
    }
    \foreach \i/\aa in {0/K,1/I,2/W,3/L,4/B,5/B} \node[font=\tiny\ttfamily, anchor=south] at (\i+0.5,-0.4) {\aa};
    \foreach \i/\w in {0/How,1/far,2/\ldots,3/About,4/12} {
      \node[font=\tiny, anchor=south west, rotate=50, inner sep=0.3pt] at (\i+6.95,-0.4) {\w};
      \node[font=\tiny, anchor=east] at (-0.15,-\i-1) {\w};
    }
    \draw[decorate, decoration={brace, amplitude=2.5pt}, rumiGreen, line width=0.5pt] (0,1.25) -- node[lab, above=2.5pt, text=rumiGreen] {protein (from folding pass)} (6,1.25);
    \draw[decorate, decoration={brace, amplitude=2.5pt}, rumiBlue, line width=0.5pt] (6.6,2.1) -- node[lab, above=2.5pt, text=rumiBlue] {text so far} (11.6,2.1);
    \node[head, anchor=south] at (5.8,4.3) {(b) Question pass ($\times 1$)};
    \node[lab, anchor=north, text width=44mm] at (5.8,-5.9) {each word attends to all of the protein and to earlier words; the protein never attends to the question};
  \end{scope}
  \begin{scope}[xshift=97mm, yshift=-2mm]
    \foreach \y/\c/\t in {0/rumiGreen/residue to residue, -1.5/rumiMint/word to protein, -3/rumiBlue/word to earlier word, -4.5/{rumiGreyL!45}/masked (later word)} {
      \fill[\c] (0,\y) rectangle ++(1,-1);
      \node[lab, anchor=west, text=rumiInk] at (1.3,\y-0.5) {\t};
    }
  \end{scope}
\end{tikzpicture}
  \caption{\textbf{One trunk, three passes, two attention masks.} Rows are the token doing the
    attending and columns the token attended to. \textbf{(a)}~The two folding passes process the
    protein alone with fully bidirectional attention; the first is recycled into the second.
    \textbf{(b)}~The question pass processes text causally, and each word also attends to the keys
    and values the second folding pass computed for every residue, at every layer.}
  \label{fig:masks}
\end{figure}
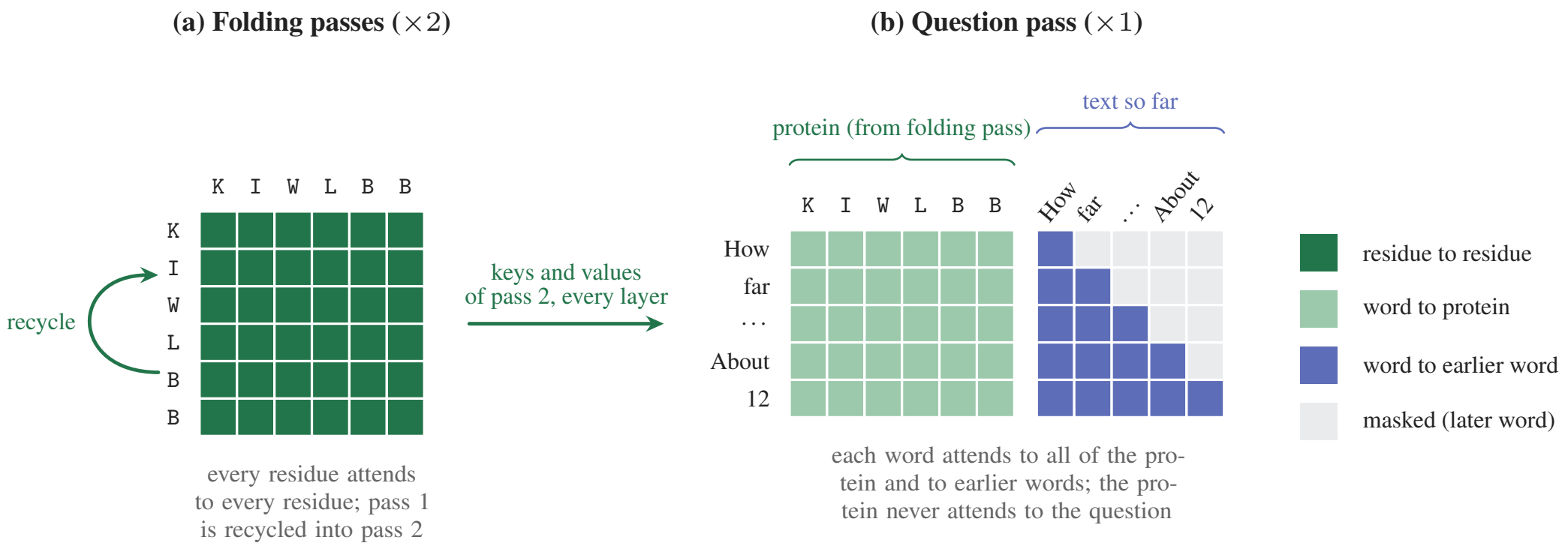

\Cref{fig:masks} shows the two attention masks used by the three passes of \cref{sec:arch}.

\subsection{Folding loss and optimisation}
\label{app:train}

\paragraph{Folding loss.}\label{app:loss}
A predicted structure should count as correct however it is rotated or shifted in space, so we
compare structures from the point of view of each residue in turn. Standing in the frame of
residue $i$, we ask where every atom of every other residue $j$ appears, in the prediction and in
the true structure (marked $^{\star}$), and measure the disagreement:
\begin{equation}
  d_{ij,a} = \big\lVert\, T_i^{-1}\mathbf{p}_{j,a} - T_i^{\star\,-1}\mathbf{p}^{\star}_{j,a} \,\big\rVert,
  \qquad
  \mathcal{L}_{\mathrm{frame}} = \operatorname*{mean}_{i,\,j,\,a}\ \min(d_{ij,a},\ 10\,\text{\AA}).
  \label{eq:frameloss}
\end{equation}
This is the frame-aligned error introduced with AlphaFold2 \citep{jumper2021highly}; capping it at
10\,\AA{} keeps a few badly misplaced residues from dominating.\footnote{In practice a tenth of the
uncapped error is added back so that distant errors still carry a gradient.} We compute it
separately for residue pairs within a chain and for pairs in different chains, so that the
interface of a complex is not swamped by the much larger number of within-chain pairs. Two small
terms keep the chain physically sensible: $\mathcal{L}_{\mathrm{bond}}$ penalises a bond between
consecutive residues whose length differs from the true one, and $\mathcal{L}_{\mathrm{clash}}$
penalises atoms closer than 1.5\,\AA{}. The full objective is
\begin{equation}
  \mathcal{L}_{\mathrm{fold}} = \mathcal{L}_{\mathrm{frame}}^{\mathrm{within}} + \mathcal{L}_{\mathrm{frame}}^{\mathrm{between}}
    + 0.5\,\mathcal{L}_{\mathrm{bond}} + 0.01\,\mathcal{L}_{\mathrm{clash}}.
  \label{eq:foldloss}
\end{equation}

\paragraph{Optimisation.}\label{app:optim}
AdamW, learning rate $10^{-4}$, weight decay 0.01, an
effective batch of 24 assemblies, gradient clipping at 1.0, on 8 A100 GPUs. No weight averaging
is used.

\subsection{Training data and held-out sets}
\label{app:data}

\paragraph{Folding.} The folding data are described in \cref{sec:stages}. We use all PDB entries released up to 2022.
For our AFDB training corpus, we remove every chain with at least 30\% sequence identity to
a CAMEO22 or CASP14 target, together with its cluster.

\paragraph{Functional training.} The stage-3 pool starts from the union of the training splits of
five protein--text sources (Prot-Inst-OOD, Prot2Text, UniProtQA, BioReason-Pro CAFA and GO-GPT),
638{,}317 distinct sequences. We screen it against the test sets of Prot2Text
(4{,}203 proteins) and the BioReason-Pro CAFA benchmark (8{,}528), our own structural evaluation
set (2{,}764), the VenusX validation and test splits, and every reviewed member of the 44 InterPro
families that VenusX holds out. A sequence is dropped if it matches one of these exactly or has an
MMseqs2 hit to one with at least 30\% identity and at least 80\% coverage of either sequence
(\texttt{easy-search}, sensitivity 7.5, e-value $10^{-3}$). This removes 365{,}157 sequences and
keeps 273{,}160, of which 205{,}016 have a Swiss-Prot entry to take labels from. The remainder is
split by MMseqs2 clusters at the same thresholds, and validation and test proteins with a training
hit are dropped (7{,}141), leaving 190{,}115 training, 2{,}413 validation and 5{,}347 test
proteins. Of the training proteins, 178{,}457 carry function, GO or EC annotation and 61{,}838
carry at least one residue-level label (26{,}450 with active sites, 50{,}319 with binding sites).
The GO-GPT test set was not part of this screen, which predates that evaluation; GO-GPT's training
split is one of the five sources, so on that benchmark only exact test sequences are excluded.

\paragraph{Task fine-tuning.} Each stage-4 model is trained on its benchmark's own training split
with no further identity screen. For GO-GPT and Mol-Instructions, training items whose protein
appears in the test set are removed by exact match. For Prot2Text we use the released split
unchanged (248{,}315 training, 4{,}172 validation and 4{,}203 test proteins, built by its authors
with CD-HIT at 40\% identity), in which no sequence is shared between training and test; sequences
longer than 1{,}024 residues are truncated. The Prot2Text fine-tune starts from the 95k-update
checkpoint, and rows are drawn at random with replacement, weighted from update 166k so that
repeated descriptions are sampled less often. Because none of these splits guarantees distance
from the training set, we report every function result stratified by sequence identity to it.

\subsection{Language ability after protein training}
\label{app:retention}

\begin{table}[h]
  \caption{\textbf{Change in accuracy relative to base Qwen3-8B, in percentage points.} Full test
    sets with direct answering, except where noted. \emph{Reasoning mode} is on 200-item subsets
    (intervals of about $\pm 6$ points). \emph{Replay only} is the base model trained on the
    replay objective alone for about 44k steps, with no protein data; it was scored with few-shot
    prompts (ARC-C 25, MMLU and GSM8K 5, TruthfulQA 0), so it is comparable with the other rows
    in scale only. Its TruthfulQA score varies by several points between checkpoints.}
  \label{tab:retention}
  \centering
  \small
  \begin{tabular}{@{}lcccc@{}}
    \toprule
    Model                                  & ARC-C & TruthfulQA & MMLU & GSM8K \\
    \midrule
    \ours{} (95k updates)                  & $-17.0$ & $-19.4$ & $-14.2$ & $-21.1$ \\
    \quad after Gene Ontology fine-tuning  & $-23.4$ & $-31.6$ & $-18.6$ & $-27.3$ \\
    \quad after Prot2Text fine-tuning      & $-11.9$ & $-25.0$ & $-10.4$ & $-13.2$ \\
    \ours{} (95k updates), reasoning mode  & $-14.5$ & $-21.0$ & $-23.8$ & $-14.0$ \\
    \midrule
    Replay only, no proteins               & $-4.7$ & $-3.8$ & $-7.4$ & $-8.7$ \\
    \bottomrule
  \end{tabular}
\end{table}

Task fine-tuning moves the cost in both directions: Gene Ontology fine-tuning adds to it on every
benchmark, and Prot2Text fine-tuning reduces it on all but TruthfulQA (\cref{tab:retention}).

\FloatBarrier
\section{Structural Questions}
\label{app:perturb}

\paragraph{Question types and perturbed folds.}
The structural questions of stage~2 ask for the distance between two residues, how buried a
residue is, which residues lie within 8\,\AA{} of another, which residues form long-range contacts
or an interface, and how confident the fold is in a segment. Every label is computed on the
coordinates the model itself predicted for that protein.

From late in stage~2, a quarter of the structural questions are asked about a deliberately degraded
fold: Gaussian noise is added to the \esmc{} input of the folding pass, with its scale adapted to
keep the perturbed lDDT between 0.4 and 0.7, and the labels are computed on the resulting
coordinates. This was introduced after an earlier analysis showed that some
answers were being read straight from the encoder features; it forces the answer to track the fold actually
produced rather than the fold the sequence would normally give.

\subsection{With and without a folding reader}
\label{app:structqa}

\begin{table}[h]
  \caption{\textbf{Structural question answering with and without a folding model as the protein
    reader.} Exact-answer accuracy on held-out proteins, with an earlier folding checkpoint as the
    reader. Both arms use Qwen3-8B, \esmc{} features
    and the same question set; \emph{No folding} reads \esmc{} through a fresh bridge. Questions
    that require three-dimensional geometry separate the arms; questions answerable from local
    sequence context do not.}
  \label{tab:structqa}
  \centering
  \small
  \begin{tabular}{@{}lcc@{}}
    \toprule
    Question type & No folding & Folding \\
    \midrule
    \emph{Geometry-dependent} & & \\
    \quad Distance between two residues (value)   & 0.290 & \textbf{0.704} \\
    \quad Closest residue to a given residue      & 0.045 & \textbf{0.444} \\
    \quad Contact partners of a residue           & 0.079 & \textbf{0.527} \\
    \quad Nearest of four candidates              & 0.361 & \textbf{0.828} \\
    \quad Long-range contact (yes/no)             & 0.722 & \textbf{0.888} \\
    \quad Strand pairing                          & 0.345 & \textbf{0.740} \\
    \emph{Sequence-readable} & & \\
    \quad Solvent exposure                        & 0.833 & 0.844 \\
    \quad Secondary-structure state               & 0.805 & 0.816 \\
    \quad Buried helix residues                   & \textbf{0.661} & 0.550 \\
    \midrule
    Headline (mean over types)                    & 0.484 & \textbf{0.642} \\
    \bottomrule
  \end{tabular}
\end{table}

\Cref{tab:structqa} compares two question-answering models trained on the same structural
questions for the same number of steps: one whose protein reader is a folding checkpoint, and one
that reads \esmc{} embeddings through a freshly initialised bridge with no folding pretraining.
On questions that can be answered from local sequence context, such as solvent exposure or
secondary structure, the two are indistinguishable. On questions that need the three-dimensional
arrangement they are far apart: 70\% against 29\% on the distance between two residues, 44\% against
5\% on naming the closest residue. \esmc{} embeddings contain a great deal of structural
information, but a language model attached to them does not learn to extract pairwise geometry in
this budget, whereas a language model that has been made to fold has it available.

\begin{table}[h]
  \caption{\textbf{Structural question answering on the current model, and the heads it reads
    through.} 264 questions about each model's own predicted structure on 24 held-out chains
    (144 C$\alpha$ distances, 72 nearer-segment and 48 residue-identity questions). Heads are zeroed
    only while the question is read; the fold is left intact. \emph{Shared text-matching contact
    heads} are contact heads in all eight checkpoints we examined that also match repeated tokens
    in base Qwen. Random heads are drawn from the same layers (mean of five draws).}
  \label{tab:structqaheads}
  \centering
  \small
  \begin{tabular}{@{}lcccc@{}}
    \toprule
    & \multicolumn{2}{c}{Distance} & Nearer & Residue \\
    \cmidrule(lr){2-3}
    Heads zeroed while reading & $r$ & MAE (\AA) & segment & identity \\
    \midrule
    \emph{\ours{} (95k updates)} & & & & \\
    \quad None (intact)                              & 0.93 & 2.6 & 0.92 & 0.96 \\
    \quad 7 shared text-matching contact heads       & 0.79 & 6.6 & 0.81 & 0.96 \\
    \quad 7 random heads                             & 0.93 & 2.6 & 0.91 & 0.96 \\
    \quad 19 text-matching contact heads             & 0.64 & 8.1 & 0.79 & 0.96 \\
    \quad 19 random heads                            & 0.93 & 2.7 & 0.91 & 0.96 \\
    \quad 3 shared contact heads, not text-matching  & 0.93 & 2.6 & 0.92 & 0.96 \\
    \quad 17 text-matching heads in layers 1--8      & 0.94 & 2.6 & 0.93 & 0.96 \\
    \emph{Structural-QA checkpoint (33k updates)} & & & & \\
    \quad None (intact)                              & 0.95 & 2.4 & 0.93 & 0.90 \\
    \quad 7 shared text-matching contact heads       & 0.79 & 5.9 & 0.86 & 0.94 \\
    \quad 7 random heads                             & 0.95 & 2.3 & 0.92 & 0.91 \\
    \quad 19 text-matching contact heads             & 0.67 & 6.8 & 0.78 & 0.94 \\
    \quad 19 random heads                            & 0.94 & 2.3 & 0.93 & 0.93 \\
    \quad 3 shared contact heads, not text-matching  & 0.95 & 2.2 & 0.93 & 0.90 \\
    \quad 17 text-matching heads in layers 1--8      & 0.94 & 2.6 & 0.93 & 0.90 \\
    \bottomrule
  \end{tabular}
\end{table}

\paragraph{Structural questions on the current model.}
\Cref{tab:structqaheads} repeats the evaluation on \ours{} itself and on an earlier structural-QA
checkpoint, with questions about each model's own fold. \ours{} states C$\alpha$ distances with a
mean error of 2.6\,\AA{} (correlation 0.93 with the distances in its fold), picks the nearer of
two segments 92\% of the time and names the residue at a position 96\% of the time.

The same evaluation shows which heads the geometric answers pass through. Seven heads are contact
heads in every checkpoint we examined and are also heads that, in base Qwen reading text, attend
from a token to an earlier copy of it (\cref{sec:interp}). Zeroing those seven only while
the question is read, with the fold untouched, raises the distance error of \ours{} to 6.6\,\AA{},
4.1\,\AA{} more than zeroing seven random heads from the same layers (95\% interval $[3.4, 4.8]$),
and lowers nearer-segment accuracy by 11 points ($[4, 18]$). Replacing their output with its mean
instead of zero gives the same result ($+3.7$\,\AA{}, $[3.0, 4.4]$). Residue identity does not
move: the model still reads the protein, and loses its pairwise geometry. The two contrast sets,
shared contact heads that are not text-matching and text-matching heads in early layers that
never become contact heads, leave every answer at the intact level.

\FloatBarrier
\section{Further Folding Results}
\label{app:fold}

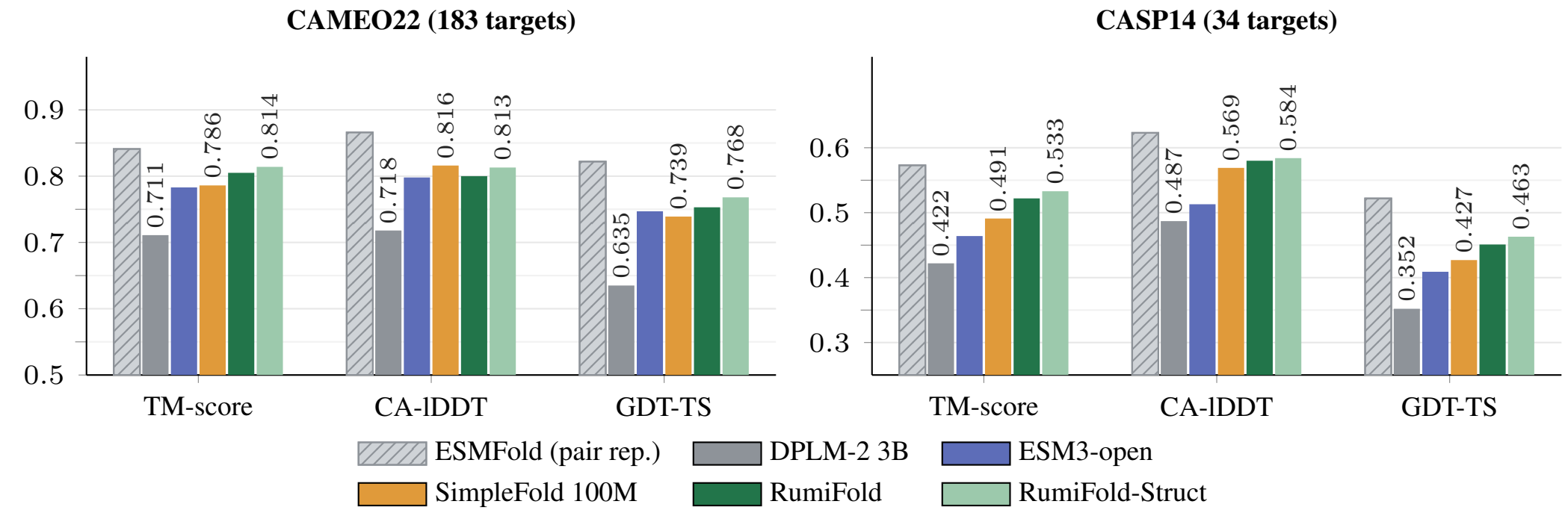
\begin{figure}[h]
  \centering
  \begin{tikzpicture}
    \begin{axis}[paper, name=cameo, ybar=0.6pt, area legend, bar width=6.6pt, width=0.55\linewidth, height=4.4cm,
        title={CAMEO22 (183 targets)}, symbolic x coords={TM,lDDT,GDT}, xtick=data,
        xticklabels={TM-score, CA-\lddt{}, GDT-TS}, enlarge x limits=0.24,
        ymin=0.5, ymax=0.98, ytick={0.5,0.6,0.7,0.8,0.9}, minor y tick num=1, yminorgrids,
        minor grid style={gray!10},
        legend to name=foldlegendb, legend columns=3]
      \addplot[reference bar] coordinates {(TM,0.841) (lDDT,0.866) (GDT,0.822)};
      \addplot[fill=cGrey, draw=none, vals=3] coordinates {(TM,0.711) (lDDT,0.718) (GDT,0.635)};
      \addplot[fill=cBaseB, draw=none] coordinates {(TM,0.783) (lDDT,0.798) (GDT,0.747)};
      \addplot[fill=cBaseA, draw=none, vals=3] coordinates {(TM,0.786) (lDDT,0.816) (GDT,0.739)};
      \addplot[fill=cOurs, draw=none] coordinates {(TM,0.805) (lDDT,0.800) (GDT,0.753)};
      \addplot[fill=cCtrl, draw=none, vals=3] coordinates {(TM,0.814) (lDDT,0.813) (GDT,0.768)};
      \legend{ESMFold (pair rep.), DPLM-2 3B, ESM3-open, SimpleFold 100M, \ours{}, \foldonly{}}
    \end{axis}
    \begin{axis}[paper, at={(cameo.south east)}, xshift=0.85cm, ybar=0.6pt, area legend, bar width=6.6pt,
        width=0.55\linewidth, height=4.4cm,
        title={CASP14 (34 targets)}, symbolic x coords={TM,lDDT,GDT}, xtick=data,
        xticklabels={TM-score, CA-\lddt{}, GDT-TS}, enlarge x limits=0.24,
        ymin=0.25, ymax=0.74, ytick={0.3,0.4,0.5,0.6}, minor y tick num=1, yminorgrids,
        minor grid style={gray!10}]
      \addplot[reference bar] coordinates {(TM,0.573) (lDDT,0.623) (GDT,0.522)};
      \addplot[fill=cGrey, draw=none, vals=3] coordinates {(TM,0.422) (lDDT,0.487) (GDT,0.352)};
      \addplot[fill=cBaseB, draw=none] coordinates {(TM,0.464) (lDDT,0.513) (GDT,0.409)};
      \addplot[fill=cBaseA, draw=none, vals=3] coordinates {(TM,0.491) (lDDT,0.569) (GDT,0.427)};
      \addplot[fill=cOurs, draw=none] coordinates {(TM,0.522) (lDDT,0.580) (GDT,0.451)};
      \addplot[fill=cCtrl, draw=none, vals=3] coordinates {(TM,0.533) (lDDT,0.584) (GDT,0.463)};
    \end{axis}
  \end{tikzpicture}\\[-2pt]
  \pgfplotslegendfromname{foldlegendb}
  \caption{Single-sequence structure prediction on the two benchmarks excluded from training at 30\%
    identity, all targets and all lengths, in the layout of \cref{fig:fold}.}
  \label{fig:fold217}
\end{figure}

\paragraph{CAMEO22 and CASP14.}
These are the two benchmarks whose targets, and everything within 30\% sequence identity of them,
were removed from our training data. We use every CAMEO22 target \citep{robin2021continuous} (183)
and every CASP14 target with public coordinates \citep{kryshtafovych2021critical} (34), with no
length cap (\cref{fig:fold217,tab:fold}). On CAMEO22 \ours{} reaches a TM-score of 0.805,
CA-\lddt{} 0.800 and GDT-TS 0.753, ahead of ESM3-open and DPLM-2 and behind SimpleFold 700M (0.816,
0.846, 0.786). On the harder targets of CASP14 it reaches 0.522, 0.580 and 0.451: ahead of
ESM3-open, DPLM-2 and SimpleFold 100M (0.491, 0.569, 0.427) and behind SimpleFold 700M (0.575,
0.632, 0.520). SimpleFold 3B is further ahead on CASP14 (0.637, 0.685, 0.589), above ESMFold
(\cref{tab:fold}). \foldonly{} is a little better than \ours{} on both (0.814 and 0.533 TM-score).

\begin{table}[h]
  \caption{Single-sequence structure prediction on all CAMEO22 (183) and CASP14 (34) targets, no
    length cap: TM-score / C$\alpha$-\lddt{} / GDT-TS, full chain, our scorer. Chai-1 and ESMFold use
    pair representations.}
  \label{tab:fold}
  \centering
  \small
  \begin{tabular}{@{}lcc@{}}
    \toprule
    Model & CAMEO22 & CASP14 \\
    \midrule
    ESMFold                 & .841 / .866 / .822 & .573 / .623 / .522 \\
    Chai-1 (single sequence) & .835 / .871 / .818 & .566 / .629 / .523 \\
    SimpleFold 3B           & .820 / .849 / .796 & .637 / .685 / .589 \\
    SimpleFold 1.1B         & .818 / .843 / .786 & .575 / .633 / .515 \\
    SimpleFold 700M         & .816 / .846 / .786 & .575 / .632 / .520 \\
    SimpleFold 100M         & .786 / .816 / .739 & .491 / .569 / .427 \\
    ESM3-open               & .783 / .798 / .747 & .464 / .513 / .409 \\
    DPLM-2 3B               & .711 / .718 / .635 & .422 / .487 / .352 \\
    DPLM-2 650M             & .672 / .676 / .589 & .385 / .460 / .320 \\
    \ours{}                 & .805 / .800 / .753 & .522 / .580 / .451 \\
    \foldonly{}             & .814 / .813 / .768 & .533 / .584 / .463 \\
    \bottomrule
  \end{tabular}
\end{table}

\begin{table}[h]
  \caption{Three recent single-chain benchmarks, scored with the benchmark authors' script:
    C$\alpha$ TM-score / \lddt{} / GDT-TS over all targets. ESMFold and Chai-1 (top block) use
    pair representations. Every model above the last rule is run from a single sequence, with no MSA
    or templates; AlphaFold3 and Boltz-2 (bottom) are given a multiple sequence alignment and are
    shown as a reference for what MSA-based prediction reaches, not as a like-for-like comparison.
    Our training data was screened only
    against CASP14 and CAMEO22, so none of these sets is guaranteed to be held out for \ours{}.}
  \label{tab:calpha}
  \centering
  \small
  \resizebox{\ifdim\width>\linewidth\linewidth\else\width\fi}{!}{%
  \begin{tabular}{@{}lccc@{}}
    \toprule
    Model & FoldBench (334) & CASP16 (155 domains) & CAMEO 2025 (200) \\
    \midrule
    ESMFold                  & .871 / .850 / .827 & .764 / .767 / .710 & .905 / .891 / .863 \\
    Chai-1                   & .871 / .853 / .827 & .758 / .770 / .707 & .921 / .910 / .885 \\
    \midrule
    SimpleFold 3B            & .840 / .807 / .780 & .706 / .727 / .647 & .936 / .919 / .902 \\
    SimpleFold 700M          & .832 / .802 / .765 & .706 / .730 / .641 & .928 / .914 / .889 \\
    SimpleFold 100M          & .796 / .765 / .711 & .662 / .702 / .589 & .897 / .877 / .835 \\
    ESM3-open                & .783 / .747 / .710 & .651 / .668 / .588 & .849 / .819 / .785 \\
    DPLM-2 3B                & .709 / .659 / .591 & .547 / .592 / .466 & .802 / .744 / .684 \\
    DPLM-2 650M              & .656 / .614 / .536 & .515 / .562 / .436 & .757 / .703 / .635 \\
    \ours{}                  & .840 / .781 / .756 & .714 / .706 / .636 & .893 / .833 / .818 \\
    \foldonly{}              & .846 / .792 / .767 & .720 / .714 / .642 & .896 / .848 / .830 \\
    \midrule
    \emph{With an MSA} & & & \\
    AlphaFold3               & .937 / .927 / .914 & .858 / .865 / .818 & .960 / .948 / .932 \\
    Boltz-2                  & .935 / .927 / .909 & .846 / .857 / .801 & .961 / .953 / .938 \\
    \bottomrule
  \end{tabular}}
\end{table}

On CAMEO 2025, the third of these sets, SimpleFold 3B is ahead of \ours{} on every measure (0.936
against 0.893 TM-score).

\paragraph{Folding against the other skills.}
\ours{} was finished with a phase of 30k updates weighted toward structure (88\% structure items).
It raised CAMEO22 \lddt{} from 0.795 to 0.806 and CASP14 from 0.569 to 0.574 on the trainer's
panel, and improved structural question answering (0.81 to 0.87), while the function measures
fell: the GO head's \fmax{} from 0.450 to 0.416 and active-site AUPR from 0.848 to 0.736. The skills
share one set of adapters and compete for it.

\begin{figure}[h]
  \centering
  \begin{subfigure}[b]{0.47\linewidth}
    \centering
    \includegraphics[width=\linewidth]{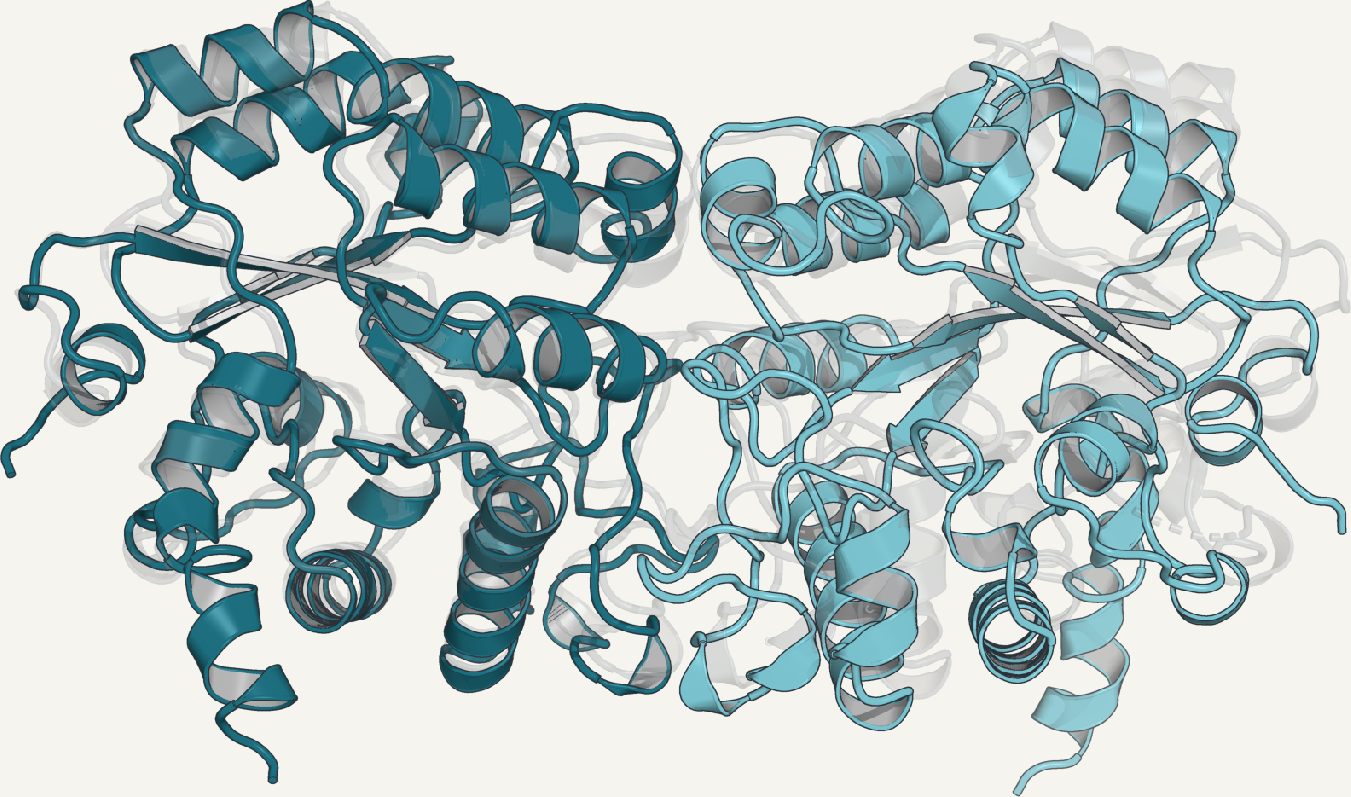}
    \caption{\foldonly{}: DockQ 0.77, interface RMSD 1.7\,\AA}
  \end{subfigure}\hfill
  \begin{subfigure}[b]{0.47\linewidth}
    \centering
    \includegraphics[width=\linewidth]{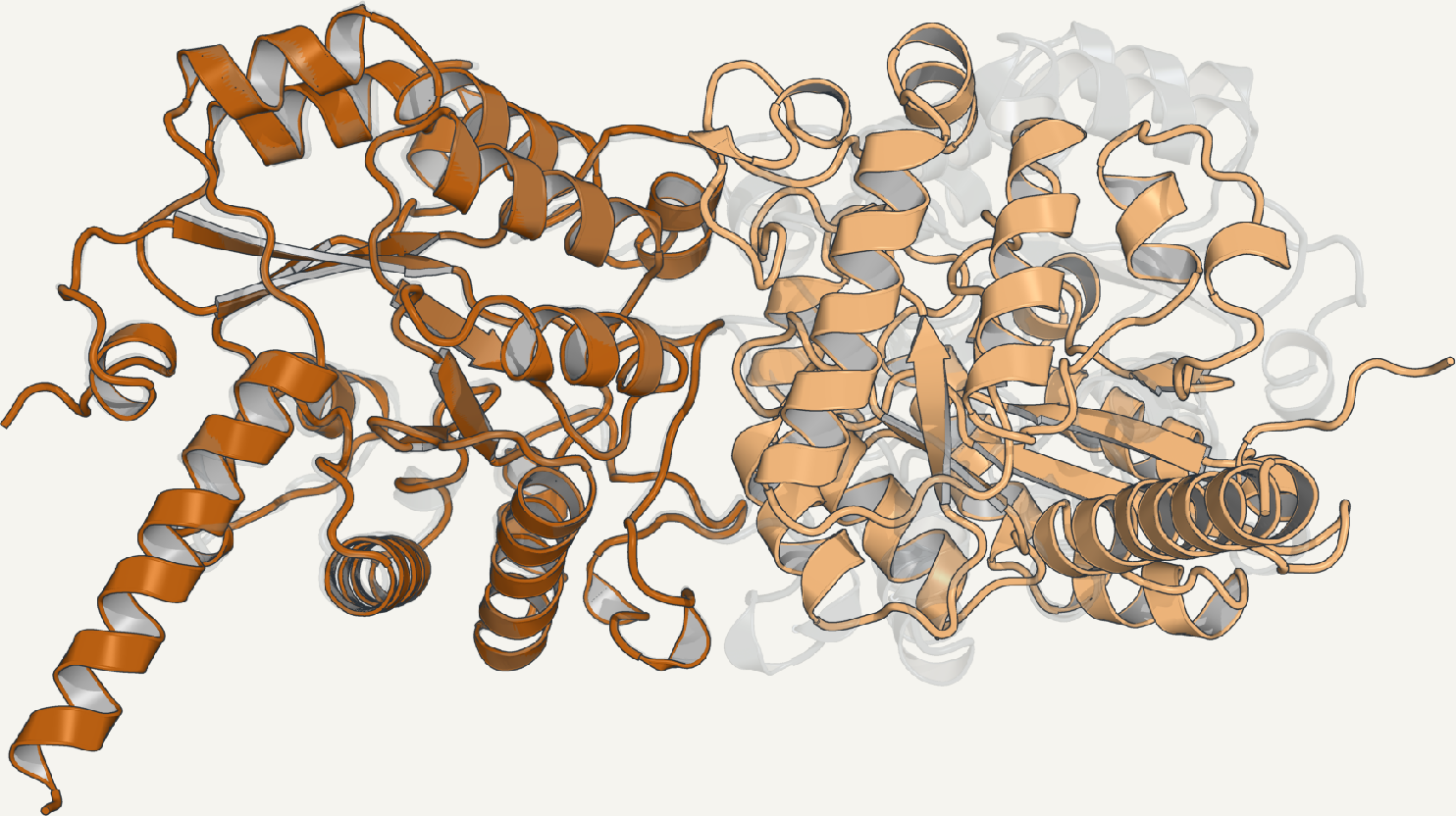}
    \caption{ESMFold: DockQ 0.04, interface RMSD 12.3\,\AA}
  \end{subfigure}
  \caption{\textbf{A held-out homodimer (PDB 8JQF, $2 \times 305$ residues).} Native structure in
    grey; each prediction is superposed on its left chain, so the right chain shows where the
    partner was placed. ESMFold folds each chain more accurately (\lddt{} 0.96 vs.\ 0.83) but docks
    it incorrectly. The closest training sequence is 40\% identical.}
  \label{fig:8jqf}
\end{figure}

\paragraph{Complexes.}\label{app:complex}
\ours{} accepts two chains and so predicts complexes directly; ESMFold can do so through a
poly-glycine linker. The results in this section are for \foldonly{}. \Cref{tab:dock} reports the fraction of FoldBench interfaces predicted
acceptably (DockQ $\geq 0.23$ \citep{basu2016dockq}), and \cref{fig:8jqf} shows an example.

\begin{table}[h]
  \caption{Docking success (DockQ $\geq 0.23$) on FoldBench interfaces. Held out: every chain below
    50\% identity to \foldonly{} training chains.}
  \label{tab:dock}
  \centering
  \small
  \begin{tabular}{@{}lrcc@{}}
    \toprule
    Interface type & $n$ & \foldonly{} & ESMFold (linker) \\
    \midrule
    Protein--protein            & 244 & \textbf{43\%} & 19\% \\
    Protein--protein, held out  &  50 & \textbf{32\%} & 26\% \\
    Antibody--antigen           & 153 & 1\%           & 0\%  \\
    Protein--peptide            &  36 & \textbf{19\%} & 14\% \\
    \bottomrule
  \end{tabular}
\end{table}

\subsection{What does the language model bring to folding?}
\label{sec:lm}

\ours{} starts from a model pretrained on text. Does that pretraining matter for folding, or would
any transformer of this size do? We ran controlled comparisons at small scale (about 2k training
proteins, two seeds), and the answer depends on the regime.

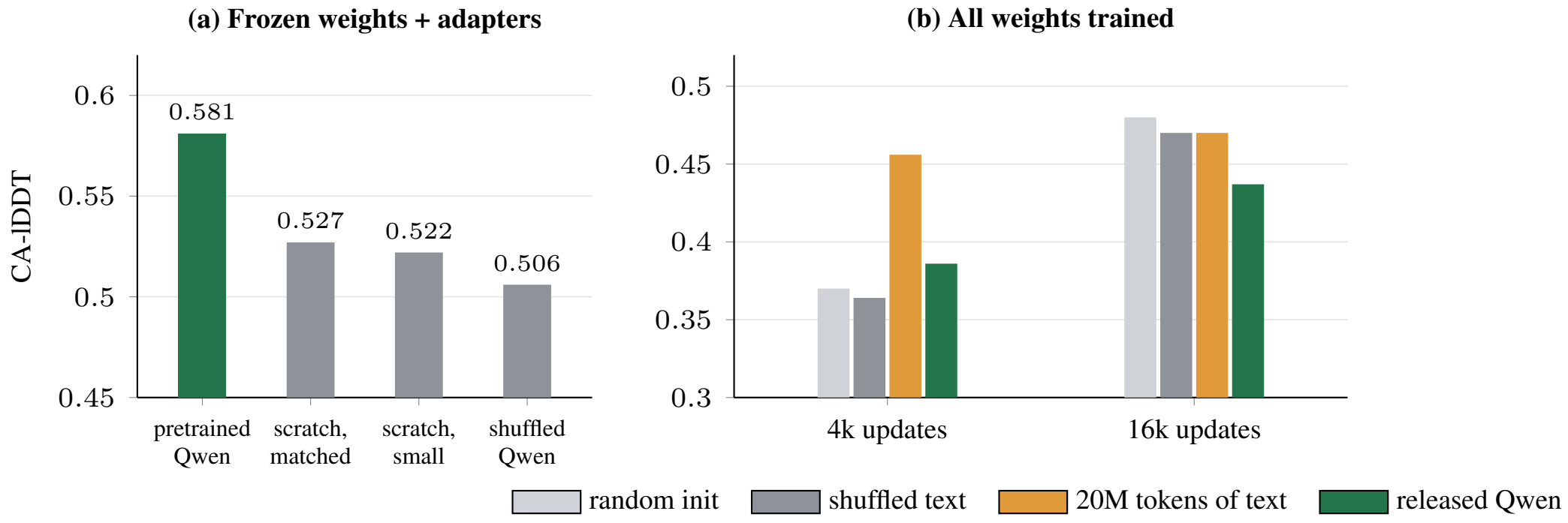
\begin{figure}[h]
  \centering
  \begin{tikzpicture}
    \begin{axis}[paper, name=frozen, ybar, bar width=12pt, width=0.40\linewidth, height=4.6cm,
        title={(a) Frozen weights + adapters},
        symbolic x coords={pre,matched,small,shuf}, xtick={pre,matched,small,shuf},
        xticklabels={{pretrained\\Qwen}, {scratch,\\matched}, {scratch,\\small}, {shuffled\\Qwen}},
        x tick label style={font=\tiny, align=center},
        enlarge x limits=0.2, ymin=0.45, ymax=0.62, ytick={0.45,0.5,0.55,0.6}, ylabel={CA-\lddt{}},
        nodes near coords, every node near coord/.append style={font=\tiny, /pgf/number format/.cd, fixed, fixed zerofill, precision=3}]
      \addplot[fill=cGrey, draw=none, bar shift=0pt] coordinates {(matched,0.527) (small,0.522) (shuf,0.506)};
      \addplot[fill=cOurs, draw=none, bar shift=0pt] coordinates {(pre,0.581)};
    \end{axis}
    \begin{axis}[paper, at={(frozen.south east)}, xshift=1.25cm, ybar=1pt, area legend, bar width=8pt,
        width=0.50\linewidth, height=4.6cm, title={(b) All weights trained},
        symbolic x coords={a,b}, xtick=data, xticklabels={4k updates, 16k updates},
        enlarge x limits=0.5, ymin=0.3, ymax=0.52, ytick={0.3,0.35,0.4,0.45,0.5},
        legend style={at={(0.5,-0.22)}, anchor=north}, legend columns=4]
      \addplot[fill=cGreyL, draw=none] coordinates {(a,0.370) (b,0.480)};
      \addplot[fill=cGrey,  draw=none] coordinates {(a,0.364) (b,0.470)};
      \addplot[fill=cBaseA, draw=none] coordinates {(a,0.456) (b,0.470)};
      \addplot[fill=cOurs,  draw=none] coordinates {(a,0.386) (b,0.437)};
      \legend{random init, shuffled text, 20M tokens of text, released Qwen}
    \end{axis}
  \end{tikzpicture}
  \caption{\textbf{Text pretraining helps when the base model is frozen, and only early when it is
    not.} Small-scale folding runs, two seeds each, development proteins.
    \textbf{(a)}~Our recipe (frozen base, rank-64 adapters), after 4k updates: pretrained Qwen
    against trunks trained from scratch and against Qwen with each weight matrix shuffled.
    \textbf{(b)}~The Qwen architecture with every weight trained, from four initialisations.
    Pretraining on text gives a head start that is gone by 16k updates. Bars are means of two
    seeds; per-seed values and intervals are in the text.}
  \label{fig:pretrain}
\end{figure}

With the base model frozen, the regime \ours{} uses, pretrained weights help. After 4k updates
pretrained Qwen with low-rank adapters is ahead of a transformer trained from scratch with the
same number of trainable parameters by 0.054 \lddt{} $[0.042, 0.069]$, and ahead of Qwen with
every frozen weight matrix randomly permuted by 0.075 $[0.058, 0.093]$
(\cref{fig:pretrain}a). The size of the effect varies between the two seeds, so we read this as
support for the direction and not for the size.

With every weight trained, the benefit is only a head start. Text pretraining speeds early
learning by 8 to 9 \lddt{} points at 4k updates, but by 16k the advantage is gone, and the
released Qwen weights finish below random initialisation in both seeds (\cref{fig:pretrain}b).
What a pretrained language model offers folding is an organisation of weights that adapters can
exploit, not knowledge that survives being trained over.

\FloatBarrier
\section{Contact Heads Carry the Fold}
\label{sec:interp}

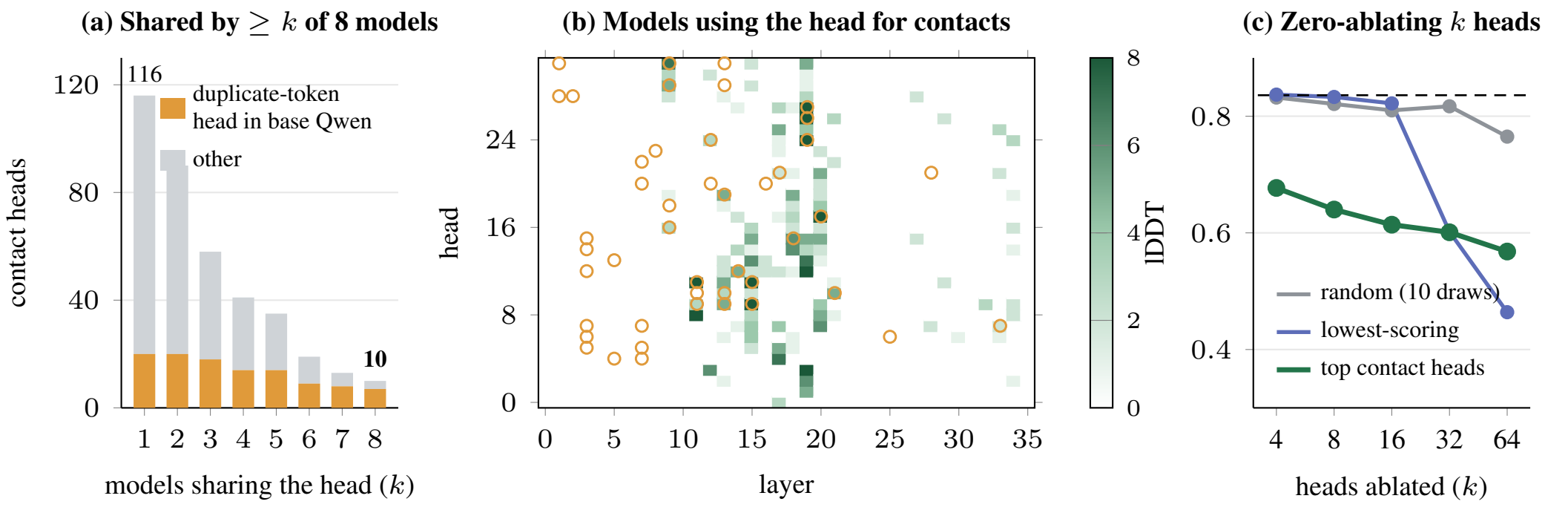
\begin{figure}[h]
  \centering
  \begin{tikzpicture}
    \begin{axis}[paper, name=share, ybar stacked, bar width=5.5pt, width=0.29\linewidth, height=4.7cm,
        legend image code/.code={\draw[#1, draw=none] (0cm,-0.08cm) rectangle (0.22cm,0.1cm);},
        title={(a) Shared by $\geq k$ of 8 models},
        xlabel={models sharing the head ($k$)}, ylabel={contact heads},
        xtick={1,2,3,4,5,6,7,8}, xmin=0.3, xmax=8.7, ymin=0, ymax=130, ytick={0,40,80,120},
        legend style={at={(0.98,0.97)}, anchor=north east, font=\tiny, cells={align=left}}, legend columns=1]
      \addplot[fill=cBaseA, draw=none] table[x=k, y=dup] {figures/draft/data/contact_heads_shared_k.dat};
      \addplot[fill=cGreyL, draw=none] table[x=k, y=other] {figures/draft/data/contact_heads_shared_k.dat};
      \legend{{duplicate-token\\head in base Qwen}, other}
      \node[font=\tiny, anchor=south] at (axis cs:1,116) {116};
      \node[font=\tiny, anchor=south] at (axis cs:8,10) {\textbf{10}};
    \end{axis}
    \begin{axis}[name=heat, at={(share.south east)}, xshift=1.25cm, width=0.43\linewidth, height=4.7cm, enlargelimits=false,
        xmin=-0.5, xmax=35.5, ymin=-0.5, ymax=31.5, unit vector ratio={},
        axis on top, tick align=outside, major tick length=2pt,
        tick label style={font=\scriptsize}, label style={font=\scriptsize},
        title style={font=\scriptsize\bfseries, yshift=-4pt},
        title={(b) Models using the head for contacts},
        xlabel={layer}, ylabel={head}, xtick={0,5,...,35}, ytick={0,8,...,31},
        colormap name=rumi,
        point meta min=0, point meta max=8,
        colorbar, colorbar style={width=2mm, ytick={0,2,4,6,8}, tick label style={font=\tiny}}]
      \addplot[matrix plot*, point meta=explicit, mesh/cols=32]
        table[x=layer, y=head, meta=n] {figures/draft/data/contact_heads_shared_count.dat};
      \addplot[only marks, mark=o, mark size=1.5pt, line width=0.55pt, draw=cBaseA]
        table[x=layer, y=head] {figures/draft/data/contact_heads_dup.dat};
    \end{axis}
    \begin{axis}[paper, at={(heat.south east)}, xshift=1.95cm, width=0.29\linewidth, height=4.7cm,
        title={(c) Zero-ablating $k$ heads},
        xlabel={heads ablated ($k$)}, ylabel={\lddt{}}, xmode=log, log basis x=2,
        xtick={4,8,16,32,64}, xticklabels={4,8,16,32,64}, ymin=0.3, ymax=0.9, ytick={0.4,0.6,0.8},
        legend style={at={(0.03,0.03)}, anchor=south west, font=\tiny},
        legend image code/.code={\draw[#1] (0cm,0cm) -- (0.3cm,0cm);}]
      \addplot[cGrey, line width=1pt, mark=*, mark size=1.3pt]
        table[x=k, y=rand] {figures/draft/data/contact_heads_ablation.dat};
      \addplot[cBaseB, line width=1pt, mark=*, mark size=1.3pt]
        table[x=k, y=bottom] {figures/draft/data/contact_heads_ablation.dat};
      \addplot[cOurs, line width=1.3pt, mark=*, mark size=1.6pt]
        table[x=k, y=top] {figures/draft/data/contact_heads_ablation.dat};
      \draw[black, densely dashed, line width=0.5pt] (axis cs:3.2,0.836) -- (axis cs:80,0.836);
      \legend{{random (10 draws)}, lowest-scoring, top contact heads}
    \end{axis}
  \end{tikzpicture}
  \caption{\textbf{Separately trained models use the same heads for contacts, and most of those
    heads match repeated tokens in base Qwen.}
    \textbf{(a)}~Number of attention heads that are contact heads in at least $k$ of eight
    checkpoints (five from the lineage of \ours{}, three trained separately). Ten are contact
    heads in all eight, and seven of the ten are duplicate-token heads in base Qwen (amber).
    \textbf{(b)}~For each of the $36 \times 32$ heads, the number of checkpoints in which it is a
    contact head; rings mark the 45 heads that, in base Qwen reading text, attend from a token to
    an earlier copy of it.
    \textbf{(c)}~\lddt{} of \foldonly{} after zeroing the output of $k$ heads (dashed: intact,
    0.836). Removing the four best contact heads costs 0.16 \lddt{}; removing four random heads
    costs nothing.}
  \label{fig:contact}
\end{figure}

We score every attention head of \foldonly{} by how well its folding-pass attention map matches the
native contact map (C$\alpha$ within 8\,\AA{}) on 24 held-out CATH chains of up to 160 residues,
using the AUC of attention weight against contacts over all pairs at least six residues apart, and
the precision of the head's top-$L$ pairs among those at least 24 apart.

\paragraph{Contact heads emerge.}
After folding training, 39 heads have contact AUC above 0.85, concentrated in layers 9 to 20 with a
second group in layers 33 and 34. With the adapters switched off, the same
network on the same inputs has no such head (maximum AUC 0.67). This echoes the contact heads of
protein language models \citep{rao2020transformer}, with the difference that Qwen was never
pretrained on proteins: the behaviour is installed entirely by low-rank updates.

\paragraph{They are necessary.}
Zeroing the four highest-scoring contact heads lowers \lddt{} from 0.836 to 0.677
(\cref{fig:contact}c). Zeroing four random heads leaves it at 0.832 on average, and none of ten
random draws falls below 0.805. The same holds for \ours{} (0.826 to 0.444) and for its Gene
Ontology fine-tune (0.851 to 0.795). At 32 or more heads the ordering reverses: removing the
lowest-scoring heads is as bad as removing the top ones or worse, so many heads matter for folding
in ways the contact score does not capture.

\paragraph{Borrowed from text.}
Where do contact heads come from in a model pretrained on text? We compared eight checkpoints,
five from the lineage of \ours{} and three trained separately from a fresh initialisation. Each
has 29 to 75 contact heads, and they are largely the same heads: ten are contact heads in all
eight (\cref{fig:contact}a,b). Seven of those ten are heads that, in base Qwen reading text, attend from a token to an
earlier copy of the same token, where chance would give 0.4. Heads that attend to the previous
token never become contact heads. The seven are needed. Zeroing them costs 0.18 \lddt{} in the
models trained only to fold, including the separately trained one, and 0.03 to 0.04 in \ours{}
and its fine-tunes, where seven random heads from the same layers cost at most 0.002. They also
carry structural question answering (\cref{tab:structqaheads}). Folding fine-tuning appears to
repurpose heads that the base model uses to find an earlier copy of a token: ``find the same
token'' becomes ``find the residue in contact''.

\FloatBarrier
\section{Further Function Benchmarks}
\label{app:brp}

\paragraph{The BioReason-Pro benchmark.}
The Gene Ontology results in \cref{sec:function} use the GO-GPT v2 benchmark. BioReason-Pro
\citep{fallahpour2026bioreason} and the first GO-GPT were trained and evaluated on an earlier
one. To compare on their terms, we train \ours{} on that benchmark's own training set and score
it on its own test set with its own scorer. Nothing else changes: the run starts from the same
95k-update checkpoint as the main model and uses the same Gene Ontology recipe, for 13 passes
over the 120{,}143 training proteins. The test set is the 8{,}159 proteins of the BioReason-Pro
release, and scores are the information-weighted F-max of that release's evaluation script,
averaged over the three aspects.

\begin{table}[h]
  \caption{\textbf{Trained on BioReason-Pro's own data, \ours{} is ahead of BioReason-Pro and
    GO-GPT on their test set.} Information-weighted F-max on the 8{,}159 test proteins, by
    aspect and averaged, with the same scorer for every row. Bold marks the
    best model in each column.}
  \label{tab:brp}
  \centering
  \small
  \begin{tabular}{@{}lcccc@{}}
    \toprule
    Model & MF & BP & CC & Mean \\
    \midrule
    BioReason-Pro (SFT) & 0.712 & 0.515 & 0.685 & 0.638 \\
    BioReason-Pro (RL)  & 0.716 & 0.543 & \textbf{0.726} & 0.662 \\
    GO-GPT v1           & \textbf{0.717} & 0.550 & 0.725 & 0.664 \\
    \ours{}, trained on the same data & 0.704 & \textbf{0.589} & \textbf{0.726} & \textbf{0.673} \\
    \bottomrule
  \end{tabular}
\end{table}

\ours{} scores 0.673, against 0.664 for GO-GPT, 0.662 for BioReason-Pro with reinforcement
learning and 0.638 for its supervised version (\cref{tab:brp}). The gain is on biological
process (0.589 against 0.550 for GO-GPT); it is level on cellular component and 0.013 behind on
molecular function.

\paragraph{Mol-Instructions.}\label{app:molinst}\label{sec:molinst}
Mol-Instructions \citep{fang2023mol} asks for a one-sentence answer about a protein under four
instructions (protein function, general function, catalytic activity, domains and motifs) and
scores it by ROUGE-L \citep{lin2004rouge} against a Swiss-Prot-derived reference. We fine-tune
\ours{} on its training items and evaluate on the full test set of 9{,}128 items, with greedy
decoding. The matched control, \emph{Qwen + \esmc{}}, is the same language model with the same
\esmc{} features, data and number of steps, but no folding pretraining.

\begin{table}[h]
  \caption{\textbf{Mol-Instructions.} ROUGE-L $\times$ 100 on all test items and on those below
    50\% and 30\% identity to the training set. \ours{}, its control and the floors are scored on
    the full test set (9{,}128 items; $n{=}2{,}433$ and $816$ in the two subsets). Published
    systems ($^\dagger$) were run on a sample of it: the 1{,}790 items that all of them answered
    ($n{=}465$ and $146$); on that sample \ours{} scores 65.6 / 55.5 / 45.3. \emph{Floor} rows
    are not trained models. Bold marks the best score in each column and underline the second.}
  \label{tab:molinst}
  \centering
  \small
  \begin{tabular}{@{}lccc@{}}
    \toprule
    System & All & $<$50\% & $<$30\% \\
    \midrule
    \ours{}                                         & \underline{66.2} & \underline{56.4} & \textbf{45.9} \\
    Qwen + \esmc{}, no folding                      & 63.1 & 52.0 & \underline{42.7} \\
    \midrule
    Mol-Instructions LLaMA-7B$^\dagger$ \citep{fang2023mol}   & 53.0 & 42.1 & 36.4 \\
    EvoLlama$^\dagger$ \citep{liu2024evollama}                & 48.6 & 38.9 & 33.4 \\
    Prot2Chat$^\dagger$ \citep{wang2025prot2chat}             & 35.6 & 32.6 & 27.2 \\
    TourSynbio$^\dagger$ \citep{shen2024toursynbio}           & 17.7 & 18.4 & 20.3 \\
    InstructProtein$^\dagger$ \citep{wang2023instructprotein} & 16.1 & 15.5 & 15.4 \\
    Evolla-10B$^\dagger$ \citep{zhou2025decoding}             & 13.0 & 12.2 & 10.8 \\
    BioMedGPT-10B$^\dagger$ \citep{luo2023biomedgpt}          &  9.9 &  9.9 &  7.9 \\
    \midrule
    \emph{Nearest training protein (floor)}         & \textbf{67.3} & \textbf{56.7} & 42.5 \\
    \emph{Template (floor)}                         & 40.7 & 40.4 & 38.8 \\
    \bottomrule
  \end{tabular}
\end{table}

\ours{} scores 66.2, against 63.1 for its matched control (\cref{tab:molinst}): folding
pretraining is worth 3.1 points overall $[2.8, 3.5]$, 4.4 below 50\% identity $[3.7, 5.2]$ and
3.2 below 30\% $[1.7, 4.7]$. The advantage holds for all four instructions. Among trained protein
models \ours{} is the state of the art on this benchmark: the best published system scores 53.0
on the sample we ran it on, where \ours{} scores 65.6. Copying the answer
of the nearest training protein remains a strong reference, because most test proteins have a
near-identical training neighbour: it scores 67.3 overall, level with \ours{} below 50\% identity,
and behind it below 30\% (42.5 against 45.9).

\FloatBarrier
\section{Representations After Folding}
\label{sec:refine}
\label{app:organise}
\label{app:venusx}

The controls of \cref{sec:controls} show that folding pretraining matters for the scores. Here we look at the
representations themselves. If learning to fold teaches the language model to describe a
protein, its states should be a better place to read a protein's properties from than the
\esmc{} features it was given, and function training should improve them further. We test this
with frozen states and simple readouts: nearest neighbours and small probes. Throughout, \esmc{} is read at its final layer, which is what the model is fed.

\paragraph{Function can be looked up in the states.}
\Cref{tab:layerscan} asks how well function can be looked up by nearest neighbours in each
representation space. The states of a model trained only to fold are more useful than the
features it reads (0.455 against 0.435 overall, and neighbours share a functional class 79\% of
the time against 70\%), although it has never seen a functional label. Function fine-tuning then
takes the folding-pretrained model much further (0.523), while the same fine-tuning applied to
plain Qwen leaves its states \emph{less} useful than the features it started from (0.427). A
second lookup on a larger test set agrees: for molecular function, neighbours in the fold-only
states score 0.462 \fmax{} against 0.391 in the \esmc{} features.

\begin{table}[t]
  \caption{\textbf{Where function is readable.} Nearest-neighbour GO lookup ($k{=}20$) in each
    representation space, from 15{,}000 training proteins to the test set; weighted \fmax{}.
    \esmc{} is read at its final layer; each model at its best layer. \emph{Purity}: share of a protein's 10 nearest neighbours in the same top-level
    molecular-function class.}
  \label{tab:layerscan}
  \centering
  \small
  \begin{tabular}{@{}lccc@{}}
    \toprule
    Representation & All & $<$30\% identity & Purity \\
    \midrule
    \esmc{}, final layer (what the models are fed)  & 0.435 & 0.363 & 0.70 \\
    Qwen + \esmc{}, GO fine-tuned (no fold pretraining)      & 0.427 & 0.351 & 0.73 \\
    \foldonly{} (folding only)                      & 0.455 & 0.372 & 0.79 \\
    \foldonly{}, GO fine-tuned                      & \textbf{0.523} & \textbf{0.420} & \textbf{0.90} \\
    \bottomrule
  \end{tabular}
\end{table}

\paragraph{Residue-level probes.}
The residue-level probes on VenusX are summarised in \cref{sec:enrich}; \cref{tab:venusx} gives every task and split.
It also gives the final layer of ESM-2 for reference. On these residue tasks ESM-2 is the
stronger encoder: both of its sizes are ahead of \esmc{} and of the fold-only states on every
active-site, binding-site, conserved-site and motif task, and behind the fold-only states only on
the three domain tasks.

The table also gives the control with no folding pretraining, \emph{Qwen + \esmc{}} fine-tuned
on Gene Ontology. Its states are a worse place to read residues from than the \esmc{} features
it was given, on all 17 tasks: by 0.11 to 0.30 AUPR on the active-site, binding-site,
conserved-site and motif tasks of the MF50 and MP50 splits, and by 0.01 to 0.09 elsewhere. It is
below the fold-only states on 15 of the 17, level on cross-family conserved sites and ahead on
cross-family active sites (0.144 against 0.101). Passing \esmc{} through a language model
trained only on function loses residue-level information that the folding objective keeps.

\begin{table}[t]
  \caption{Residue-level probes on VenusX, final layer of every model: pooled test AUPR of the
    same probe fit on frozen features of each encoder, on \foldonly{}'s final second-pass
    states and on the final-layer states of \emph{Qwen + \esmc{}}, the control fine-tuned on
    Gene Ontology with no folding (\cref{sec:controls}), for each task and split (MF50, MP50
    and cross-family). Bold marks the higher of
    \esmc{} and \foldonly{}; the ESM-2 columns are for reference.
}
  \label{tab:venusx}
  \centering
  \small
  \setlength{\tabcolsep}{4pt}
  \resizebox{\ifdim\width>\linewidth\linewidth\else\width\fi}{!}{%
  \begin{tabular}{@{}lcc|ccc@{}}
    \toprule
    Task, split & ESM-2 650M & ESM-2 3B & \esmc{}-6B & Qwen + \esmc{} & \foldonly{} \\
    \midrule
    Active site, MF50                  & .873 & .860 & .707 & .488 & \textbf{.789} \\
    Active site, MP50                  & .944 & .944 & .864 & .679 & \textbf{.922} \\
    Active site, Cross-family          & .154 & .154 & \textbf{.151} & .144 & .101 \\
    \addlinespace[2pt]
    Binding site (InterPro), MF50      & .933 & .923 & \textbf{.842} & .643 & .804 \\
    Binding site (InterPro), MP50      & .966 & .966 & .918 & .761 & \textbf{.923} \\
    Binding site (InterPro), Cross-family & .292 & .327 & .252 & .167 & \textbf{.269} \\
    \addlinespace[2pt]
    Binding site (BioLiP), MP50        & .399 & .405 & \textbf{.395} & .315 & .390 \\
    \addlinespace[2pt]
    Conserved site, MF50               & .902 & .912 & .788 & .491 & \textbf{.831} \\
    Conserved site, MP50               & .908 & .922 & .792 & .505 & \textbf{.818} \\
    Conserved site, Cross-family       & .247 & .261 & \textbf{.261} & .209 & .209 \\
    \addlinespace[2pt]
    Motif, MF50                        & .899 & .917 & \textbf{.812} & .620 & .786 \\
    Motif, MP50                        & .849 & .849 & .780 & .666 & \textbf{.802} \\
    Motif, Cross-family                & .486 & .514 & .433 & .364 & \textbf{.436} \\
    \addlinespace[2pt]
    Domain, MF50                       & .613 & .615 & .578 & .502 & \textbf{.640} \\
    Domain, MP50                       & .618 & .630 & .598 & .533 & \textbf{.654} \\
    Domain, Cross-family               & .513 & .471 & .593 & .517 & \textbf{.637} \\
    \addlinespace[2pt]
    Epitope, MP50                      & .167 & .170 & \textbf{.179} & .149 & .177 \\
    \bottomrule
  \end{tabular}}
\end{table}

\paragraph{Function training refines them again.}
The same holds for free-text function. Copying the description of the nearest training protein in the
final \esmc{} layer scores 33.5 BLEU-4 on the Prot2Text test set; the same lookup in the states
of \ours{} after Prot2Text training, at a depth chosen without the test set, scores 44.6. The
input to the same model's first layer scores 37.7, so most of the gain is made inside the
language model.

\subsection{Structure in the hidden states}
\label{app:influence}

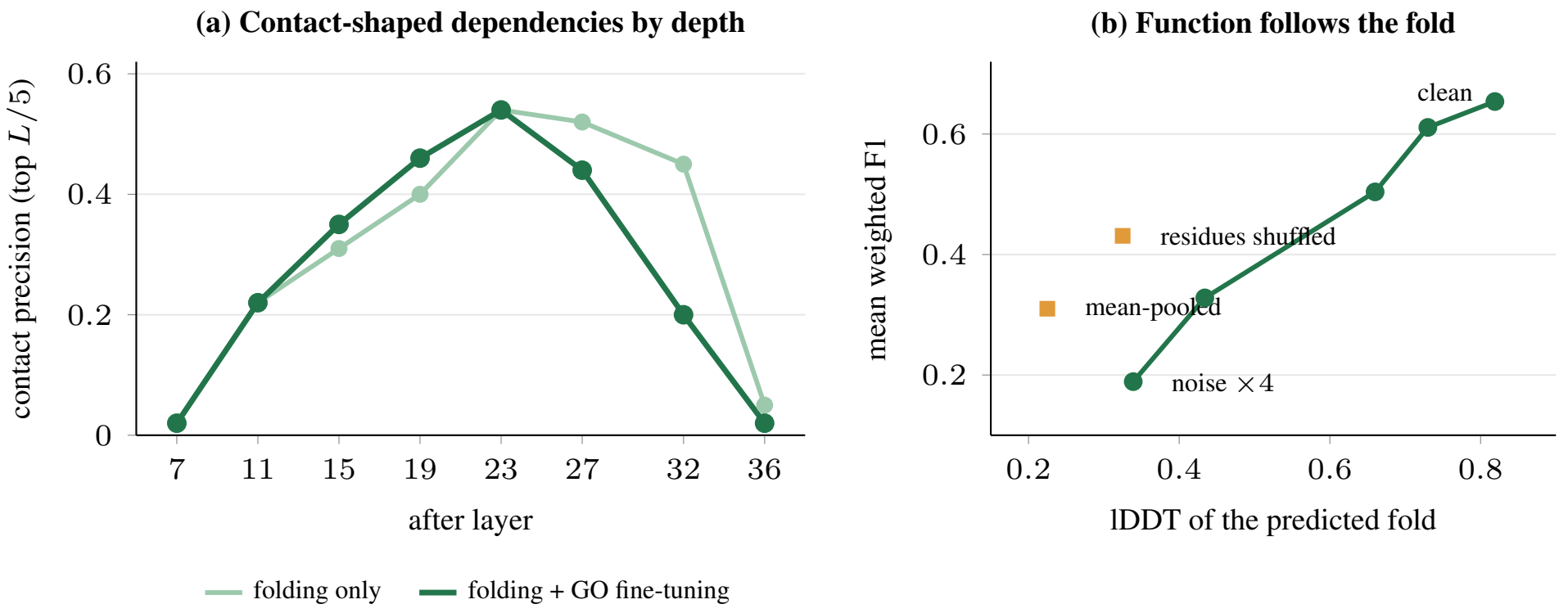
\begin{figure}[h]
  \centering
  \begin{tikzpicture}
    \begin{axis}[paper, name=depth, width=0.5\linewidth, height=4.6cm,
        title={(a) Contact-shaped dependencies by depth},
        xlabel={after layer}, ylabel={contact precision (top $L/5$)},
        xmin=5, xmax=38, xtick={7,11,15,19,23,27,32,36}, ymin=0, ymax=0.62, ytick={0,0.2,0.4,0.6},
        legend style={at={(0.5,-0.34)}, anchor=north, font=\tiny}, legend columns=2,
        legend image code/.code={\draw[#1] (0cm,0cm) -- (0.3cm,0cm);}]
      \addplot[cCtrl, line width=1.1pt, mark=*, mark size=1.4pt]
        coordinates {(7,0.02) (11,0.22) (15,0.31) (19,0.40) (23,0.54) (27,0.52) (32,0.45) (36,0.05)};
      \addplot[cOurs, line width=1.3pt, mark=*, mark size=1.6pt]
        coordinates {(7,0.02) (11,0.22) (15,0.35) (19,0.46) (23,0.54) (27,0.44) (32,0.20) (36,0.02)};
      \legend{folding only, folding + GO fine-tuning}
    \end{axis}
    \begin{axis}[paper, at={(depth.south east)}, xshift=1.5cm, width=0.44\linewidth, height=4.6cm,
        title={(b) Function follows the fold},
        xlabel={\lddt{} of the predicted fold}, ylabel={mean weighted F1},
        xmin=0.15, xmax=0.9, xtick={0.2,0.4,0.6,0.8}, ymin=0.1, ymax=0.72, ytick={0.2,0.4,0.6}]
      \addplot[cOurs, line width=1.1pt, mark=*, mark size=1.6pt]
        coordinates {(0.819,0.654) (0.730,0.611) (0.660,0.504) (0.434,0.328) (0.339,0.189)};
      \addplot[only marks, mark=square*, mark size=1.6pt, cBaseA] coordinates {(0.325,0.431) (0.225,0.310)};
      \node[font=\tiny, anchor=west] at (axis cs:0.355,0.431) {residues shuffled};
      \node[font=\tiny, anchor=west] at (axis cs:0.255,0.31) {mean-pooled};
      \node[font=\tiny, anchor=east] at (axis cs:0.81,0.67) {clean};
      \node[font=\tiny, anchor=west] at (axis cs:0.37,0.185) {noise $\times 4$};
    \end{axis}
  \end{tikzpicture}
  \caption{\textbf{Structure inside the language model, and its link to function.}
    \textbf{(a)}~Each residue of the input is blanked in turn and the change in every other
    residue's state recorded; the resulting map is compared with the true contact map (40
    proteins; chance is 0.016). Contact-shaped dependencies appear by layer 11 and peak at layer
    23, and function fine-tuning leaves that peak intact.
    \textbf{(b)}~Degrading the input to the language model lowers the accuracy of the fold and of
    the function prediction together (\foldonly{} + GO head; line: one pass only, then increasing
    noise).}
  \label{fig:influence}
\end{figure}

\paragraph{Contacts in the dependencies.}
We ask directly whether the states are structured like a protein. For 40 test proteins
we blank each residue in turn and record how much every other residue's state changes. The
resulting map of who-depends-on-whom recovers the protein's contact map: among the strongest
long-range dependencies after layer 23, 54\% are true contacts, against a base rate of 1.6\%
(\cref{fig:influence}a). Function fine-tuning leaves this intact through layer 23 and changes only
the later layers, where the states become more responsive and less contact-shaped. The middle of
the network holds the structure; the end of it is rewritten for the task.

\paragraph{Function follows the fold.}
Damaging the fold damages the function prediction. As the input to the language model is
degraded, \lddt{} and weighted F1 fall together, from 0.82 and 0.65 to 0.34 and 0.19
(\cref{fig:influence}b). Shuffling the residues destroys the fold (0.33) but leaves more of the
function (0.43), as expected if part of the function signal is composition and the rest is
arrangement.

\subsection{The enrichment is computed during folding}
\label{app:during}

We limit how far along the chain each residue can see during the folding passes, leaving the
encoder's features untouched, and ask the model about a function its protein has. When no residue
sees another, it recognises the function for 18\% of proteins; the answer returns as the reach
grows, to 62\% at eight positions, 85\% at thirty-two and 100\% with the whole chain. It is also built up through the depth of the pass: with the states of an
earlier layer handed to the text, molecular-function F1 is 0.39 at the input, 0.55 at layer 24
and 0.67 for the full model. And folding links the residues of an active site: disturbing the
input at one catalytic residue changes what is stored at its partner, typically 145 positions
away, twice as much as at other residues the same distance away in space (0.17 against 0.08;
\cref{app:e2e}). Function is not passed through from the encoder. It is worked out while the
model folds.

\FloatBarrier
\section{The Folding Workspace}
\label{app:e2e}

This section gives the full results behind \cref{fig:workspace}b, and walks through individual
enzymes. The model is \ours{} after GO training (step 187{,}550). Catalytic residues are UniProt
active-site annotations with two residues on either side, so nothing about them is fitted to the
model. The question is ``Does this protein have the GO molecular-function term `catalytic
activity'?'', and the proteins are held-out enzymes that answer yes when unedited.

\begin{table}[h]
  \caption{\textbf{Mutate, restore, block: all conditions.} Share of held-out enzymes that say
    they are catalytic, on the 1{,}037 of the 1{,}448 enzymes in \cref{fig:workspace}b that have 80 to 500 residues, where every condition was run (brackets are 95\% bootstrap
    intervals). The last column is restricted to the 339 enzymes whose answer the mutation changed. \emph{Readers} are the
    attention heads that look most at the catalytic residues of that protein, chosen without
    labels; a blocked head is given what it read from the mutated protein.}
  \label{tab:e2e}
  \centering
  \small
  \begin{tabular}{@{}lcc@{}}
    \toprule
    Condition & All enzymes (\%) & Changed by the mutation (\%) \\
    \midrule
    Unedited & 100 & 100 \\
    \midrule
    \textbf{1.} Catalytic residues mutated to alanine & 67 [64, 70] & 0 \\
    \quad same number of random residues mutated & 97 [96, 98] & 93 \\
    \midrule
    \textbf{2.} Mutated, memory restored at the catalytic residues & 96 [95, 97] & 90 \\
    \quad memory restored at other residues & 67 [65, 70] & 3 \\
    \quad values restored, keys left mutated & 88 [86, 90] & 64 \\
    \quad keys restored, values left mutated & 73 [70, 76] & 26 \\
    \midrule
    \textbf{3.} Restored, then 32 readers blocked & 86 & 57 \\
    \quad 64 readers blocked & 79 & 36 \\
    \quad 128 readers blocked & 74 & 22 \\
    \quad 256 readers blocked & 70 & 11 \\
    \quad 256 random heads blocked & 93 & 80 \\
    \bottomrule
  \end{tabular}
\end{table}

\paragraph{All conditions.}
\Cref{tab:e2e} lists every condition. The memory of a residue has two parts at each layer: a key,
which decides which heads look at it, and a value, which is what they read. Restoring both brings
back 90\% of the answers the mutation changed; restoring the values alone brings back 64\%, and
the keys alone 26\%. Blocking is graded: 32 of the 1{,}152 heads undo about a third of the
restored answers, and 256 undo nearly nine in ten, where 256 random heads undo about one in ten.

\begin{figure}[h]
  \centering
  \includegraphics[width=\linewidth]{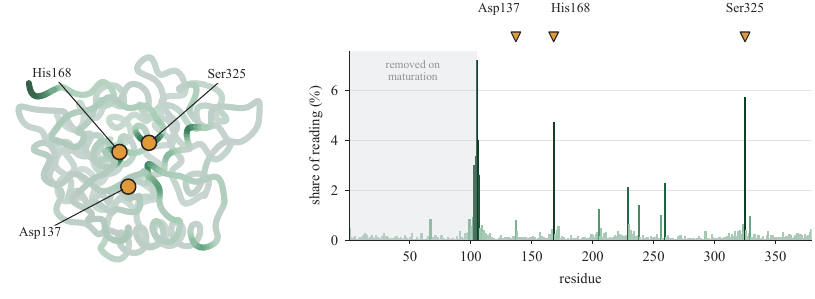}
  \caption{\textbf{Subtilisin Carlsberg (P00780).} \emph{Left:} the mature enzyme, coloured by how
    much the answer reads from each residue, with the catalytic triad marked. \emph{Right:} the
    share of the answer's attention on each residue of the 379-residue precursor. His168 and
    Ser325 are the third and second most-read residues; the most-read stretch, residues 103 to
    107, is where the propeptide is cut from the mature enzyme. The structure is the AlphaFold
    Database model, used for display only.}
  \label{fig:example}
\end{figure}

\begin{table}[h]
  \caption{\textbf{Four enzymes, step by step.} The model's written answer to ``What are this
    protein's molecular functions?'' for the unedited protein and after the catalytic residues
    and their neighbours are mutated to alanine. In all four, restoring the memory at those
    residues returns the unedited answer, mutating random residues leaves it unchanged, and
    blocking 256 readers turns the catalytic-activity answer back to no. These are chosen as
    clean cases.}
  \label{tab:examples}
  \centering
  \small
  \begin{tabular}{@{}>{\raggedright\arraybackslash}p{3.3cm}>{\raggedright\arraybackslash}p{4.8cm}>{\raggedright\arraybackslash}p{4.8cm}@{}}
    \toprule
    Enzyme (catalytic residues) & Unedited & Catalytic residues mutated \\
    \midrule
    Subtilisin Carlsberg (Asp137, His168, Ser325) &
      protein binding; serine-type endopeptidase activity & protein binding \\
    \addlinespace
    tRNA adenosine deaminase ADAT2 (Glu73) &
      tRNA-specific adenosine-34 deaminase activity; protein binding & protein binding \\
    \addlinespace
    Dual specificity phosphatase 21 (Cys105) &
      protein tyrosine phosphatase activity; protein binding & protein binding \\
    \addlinespace
    Aldose 1-epimerase (His205, Glu345) &
      aldose 1-epimerase activity & aldose binding; beta-N-acetylglucosaminidase activity \\
    \bottomrule
  \end{tabular}
\end{table}

\paragraph{Worked examples.}
\Cref{fig:example} shows where the answer reads from in subtilisin, and \cref{tab:examples}
follows four enzymes through the steps. In each, mutating the catalytic residues removes the
enzyme's own activity from the written answer, and restoring the memory at those residues brings
it back. In a sample of 264 of the enzymes, 48 follow the full pattern of yes and no answers at
every step and control, and 13 of those also change their written answer in this way.

\paragraph{Limits.}
Most enzymes keep their answer when the catalytic residues are mutated, and the written list of
functions changes less often than the yes or no answer. Single substitutions have a small effect:
of 699 substitutions recorded in UniProt as abolishing or reducing activity, each applied alone,
12\% change the model's answer, against 7\% of 475 recorded as having no effect. The model is
sensitive to these mutations in the right direction, but it does not predict the phenotype of a
point mutant.

\paragraph{Folding links the residues of an active site.}
With the input disturbed at one catalytic residue, the memory at its most distant catalytic
partner changes by 0.171 $[0.163, 0.180]$ (relative change), against 0.083 at residues matched for
distance in space. The effect peaks in the middle layers of the folding pass and is absent when
attention is limited to $\pm 8$ residues.

\subsection{Tracing function through the value memory}
\label{app:chain}

This section gives the definitions and the further results behind \cref{tab:chain}. All of it uses
\ours{} after GO training (step 187{,}550), whose text pass has no encoder path: the per-layer
key/value rows are its only view of the protein. Every edit changes value rows or head retrievals
and leaves the keys untouched. The protocol was fixed on validation proteins and run once on test
proteins; answers are greedy.

\begin{table}[h]
  \caption{\textbf{From the value memory to the answer.} Held-out test proteins that have a GO
    term $T$ and say so (215 proteins, nine terms). \emph{Says yes}: the model answers yes when
    asked whether the protein has $T$. \emph{Writes $T$}: $T$ appears in its free answer (83
    answers). ``Wiped'' replaces every value row with the training mean. Brackets are 95\%
    bootstrap intervals.}
  \label{tab:chain}
  \centering
  \small
  \setlength{\tabcolsep}{5pt}
  \begin{tabular}{@{}llcc@{}}
    \toprule
    Value memory & Text heads & Says yes (\%) & Writes $T$ (\%) \\
    \midrule
    intact & intact & 100 & 81 \\
    wiped  & intact & 0 & 0 \\
    \midrule
    wiped, $T$ coordinate restored      & intact & 19 [13, 24] & 30 [20, 40] \\
    wiped, random coordinate restored   & intact & 0 & 0 \\
    wiped, $T$ coordinate restored      & $T$'s top 32 held at their mean & 6 [3, 9] & 10 \\
    wiped, $T$ coordinate restored      & 32 random heads held & 19 & 27 \\
    \midrule
    wiped  & $T$'s top 32 given their clean retrieval & 47 [40, 53] & 57 [46, 66] \\
    wiped  & 32 random heads given theirs & 2 & 2 \\
    \midrule
    intact & $T$'s top 32 held at their mean & 86 & 77 \\
    $T$ coordinate removed & intact & 88 & 81 \\
    \bottomrule
  \end{tabular}
\end{table}

\paragraph{Definitions.}
The direction for a term $T$ is, at each layer, the mean value row of training proteins with $T$
minus the mean over those without it, fitted on 16{,}000 training proteins. A head's retrieval is
its attention-weighted sum of value rows; a term's top 32 heads are the 32 of 1{,}152 whose
retrieval best separates $T$ on 6{,}000 training proteins. Wiping sets every residue's value row,
at every layer, to the training mean. The nine terms are molecular-function terms such as ATP
binding, calcium ion binding, DNA binding and protein kinase activity; the test pools are small
for zinc (6 proteins), heme (5) and methyltransferase activity (4).

\paragraph{One coordinate, and the heads it travels through.}
\Cref{tab:chain} gives the full chain. With the memory wiped, restoring only each row's coordinate
along $T$'s direction, in layers 8 to 24, brings the function back in part (19\% say yes, $T$
written in 30\% of answers), where a random coordinate does nothing. Holding $T$'s top 32 heads
at their mean cuts that rescue to 6\%; holding 32 random heads leaves it at 19\%. The retrieval of
the top 32 heads moves the yes/no logit directly ($+0.62$, against $+0.08$ for random heads).

\paragraph{Heads.}
With all retrieval wiped, 13\% of the 215 proteins still say yes. Keeping only a term's top 8
heads gives 44\% (13\% for 8 random heads) and its top 32 gives 65\% (19\%). Keeping only the
coordinate along $T$'s direction in those 32 heads gives 63\%, against 12\% for a random
direction. The heads that recur across terms were found independently by a decoding analysis of
an earlier checkpoint.

\paragraph{Transplants.}
Copying a donor's value rows into a recipient, matched by attention rank within each layer, moves
the written answer towards the donor's (200 test pairs with no molecular-function term in common):
F1 against the donor's answer is 0.25 unedited, 0.47 with 8 rows per layer, 0.52 with 32, 0.56
with all rows, and 0.68 when keys are copied as well; 32 random rows give 0.27. The read is spread
over depth: layers 8 to 32 alone reproduce the full effect and no 12-layer band does. The same
transplant moves free-text descriptions in the Prot2Text model (ROUGE-L against the donor's
description 0.10 to 0.30, random rows 0.10).

\paragraph{Limits.}
These are tests of sufficiency. Removal is weak: taking out $T$'s coordinate, or holding its 32
heads, leaves 88\% and 86\% of proteins still saying yes, so one class direction is one carrier
among many. No edit we tried is surgical: every direction writes or removes a term only at doses
where much of the rest of the answer changes too. Residue-level statements in the model's
reasoning (``binds zinc at Cys~14'') are not read from the rows they cite: wiping those rows keeps
the statement 93\% of the time.

\subsection{Residue interventions at the input}
\label{sec:sites}

\begin{table}[h]
  \caption{\textbf{Functional sites matter, through the folding path.} For test proteins with an
    annotated site and the matching GO term, the site's residues are blanked in the input to the
    language model only (the head's direct view of the encoder is untouched). \emph{Top:} change
    in the logit of the matching term, with the change for the same number of random residues in
    parentheses. \emph{Bottom:} for \ours{}, how often the written answer changes.}
  \label{tab:sites}
  \centering
  \small
  \resizebox{\ifdim\width>\linewidth\linewidth\else\width\fi}{!}{%
  \begin{tabular}{@{}lcc@{}}
    \toprule
    \emph{Logit of the matching GO term} & \foldonly{} + GO head & Qwen + \esmc{} \\
    \midrule
    Active site $\rightarrow$ catalytic activity ($n{=}145$)      & $-0.95$ ($+0.01$) & $-0.001$ \\
    Metal ligands $\rightarrow$ metal ion binding ($n{=}35$)      & $-0.99$ ($-0.02$) & $-0.05$ \\
    Transmembrane segments $\rightarrow$ membrane ($n{=}210$)     & $-4.08$ ($-0.76$) & $-0.22$ \\
    \midrule
    \emph{Written answer of \ours{}} & Answer changed & Function no longer written \\
    \midrule
    Active site, median one residue ($n{=}135$)                   & 39\% & 13\% \\
    \quad same number of random residues                          & 4\%  & 1\% \\
    Residues in 3D contact with the active site ($n{=}89$)        & 16\% & 1\% \\
    \quad matched residues far from it                            & 3\%  & 0\% \\
    DNA-binding region ($n{=}11$)                                 & 91\% & 64\% \\
    \quad random segments                                         & 32\% & 5\% \\
    \bottomrule
  \end{tabular}}
\end{table}

If the model's prediction comes from a representation of the protein's structure, it should
depend on the residues that carry out the function, and on their neighbours in space.

\paragraph{Functional sites.}
We take test proteins with an annotated functional site and the matching GO term, and blank the
site's residues in the input to the language model, leaving the head's direct view of the encoder
intact. In the folding-pretrained model, removing an active site, a median of one residue, lowers
the logit of catalytic activity by 0.95, while removing the same number of random residues does
nothing (\cref{tab:sites}). Metal ligands and transmembrane segments behave the same way. In the
control the same intervention has no effect, because its prediction does not pass through the
language model. The effect reaches the written answer: \ours{} changes its answer in 39\% of
proteins when the active site is blanked, against 4\% for random residues, and the changes are
specific. A DNA methyltransferase keeps ``DNA binding'' and loses its methyltransferase terms; a
metallocarboxypeptidase keeps ``zinc ion binding'' and loses its peptidase terms.

\paragraph{Neighbours in space.}
The more direct test of structure uses residues that are close to the active site in three
dimensions but far from it in sequence. Blanking about seven such residues changes the written
answer in 16\% of proteins, against 3\% for residues matched in burial and sequence distance but
far away in space ($p = 0.001$).

\subsection{Sparse autoencoders on the folding memory}
\label{app:sae}

The interventions in \cref{sec:workspace} edit the memory along directions we chose. Sparse
autoencoders ask what directions the memory has of its own. We train TopK autoencoders
\citep{gao2024scaling} with 16{,}384 features, 64 active at a time, on the values the language
pass reads (five layers, 8 to 24), for \ours{} before and after Gene Ontology training, and on
the raw \esmc{} input for comparison. Each is trained on 104 million residues from the training
corpus, with the Gene Ontology validation and test proteins excluded.

\paragraph{The features carry what the answer uses.}
The autoencoder for the Gene Ontology model leaves 9\% of the variance unexplained on held-out
proteins. Writing its reconstruction back into the memory in place of the original values
changes the answers little: on 300 validation proteins, the F1 of the generated terms is 0.774
against 0.781 unedited.

\paragraph{What the features are.}
We label features against Swiss-Prot annotations on 100{,}000 proteins: 562 residue-level
concepts (feature types, ligands, modifications, secondary structure) and 8{,}714 protein-level
ones (families, domains, Gene Ontology terms, enzyme classes). A feature receives a label only
if it predicts the concept on held-out proteins better than an untrained dictionary of the same
size does. By that rule 13.7\% of the features of the Gene Ontology model are labelled, against
1.1\% for the untrained dictionary (\cref{tab:sae}). Most labelled features are detectors of a
protein family or domain that fire on a few residues of each member. About one in a hundred is
specific to a kind of residue: the histidines that bind ubiquinone (F1 0.96), cysteines in
disulfide bonds (0.75), the KMSKS motif of aminoacyl-tRNA synthetases (0.71), signal peptides
(0.68) and the cysteines of iron--sulfur clusters (0.64).

\begin{table}[h]
  \caption{\textbf{Labelled features and concept coverage.} \emph{Labelled}: share of live
    features that pass the labelling rule. \emph{Coverage}: number of annotated concepts that
    have a dedicated feature, of 179 domain, repeat and motif concepts, 141 binding-site
    concepts and 69 active-site concepts.}
  \label{tab:sae}
  \centering
  \small
  \begin{tabular}{@{}lcccc@{}}
    \toprule
    Autoencoder trained on & Labelled (\%) & Domain / motif & Binding site & Active site \\
    \midrule
    \esmc{} input & 21.3 & 37 & 17 & 7 \\
    \ours{} memory, before GO training & 13.6 & 50 & 32 & 13 \\
    \ours{} memory, after GO training & 13.7 & \textbf{64} & \textbf{35} & \textbf{19} \\
    Untrained dictionary & 1.1 & -- & -- & -- \\
    \bottomrule
  \end{tabular}
\end{table}

\paragraph{Folding adds site-level features.}
The \esmc{} input has more labelled features in total, most of them family detectors. The
folding memory has more features dedicated to functional sites: 19 of 69 active-site concepts
have their own feature after Gene Ontology training and 13 before it, against 7 in \esmc{}, and
the same order holds for binding sites and for domains and motifs (\cref{tab:sae}). Most of
this is present before any Gene Ontology training.

\paragraph{Limits.}
These are descriptions and not causal tests: no feature has been ablated or added, and a missing
Swiss-Prot annotation counts as a negative, so the label scores are lower bounds.

\subsection{Case studies}
\label{app:cases}

\begin{figure}[H]
  \centering
  \includegraphics[width=\linewidth]{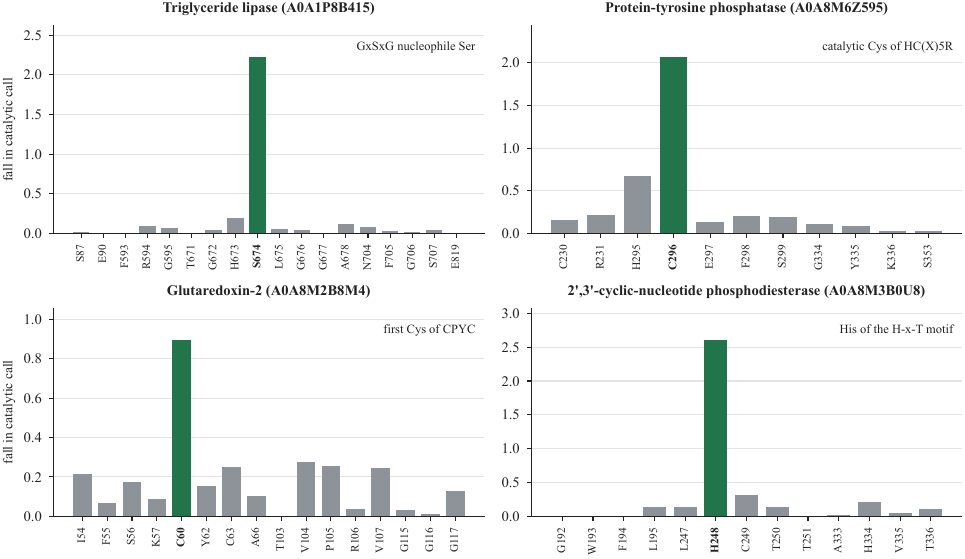}
  \caption{Single-residue knock-outs inside the most important patch, for four enzymes without
    an annotated active site. Bars show how far the model's catalytic call (a logit) falls; the
    largest is in green. The note in each panel is the textbook catalytic residue of the family.
    Final Gene Ontology checkpoint.}
  \label{fig:q-scan}
\end{figure}

\paragraph{Finding the residue a call depends on.}
\Cref{fig:q-scan} shows single-residue knock-outs on four enzymes that have no
active-site annotation in UniProt. Inside the patch of the predicted fold whose removal lowers
the model's catalytic call the most, each residue is knocked out alone. In each case one residue
carries most of the effect, and it is the residue the enzyme family's known mechanism points to.
The readout also misses: for three cGAS-like nucleotidyltransferases and an adenylate kinase among
the same twelve case studies, the decisive residues were glycines of a nucleotide-binding loop
and not the known catalytic residues.

\begin{figure}[H]
  \centering
  \includegraphics[width=\linewidth]{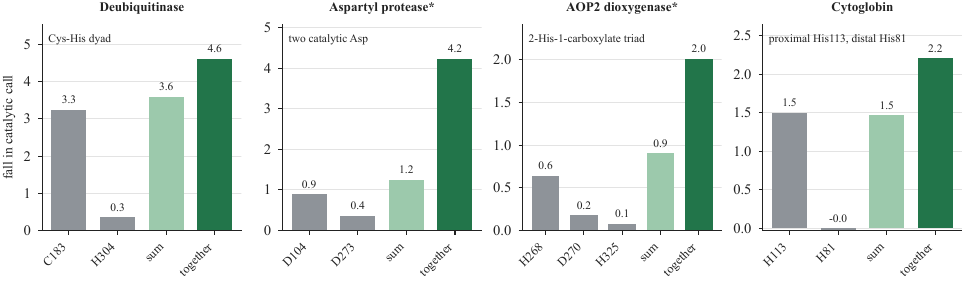}
  \caption{Residues knocked out alone (grey), the sum of those effects (light green) and the
    effect of knocking them out together (dark green). $^*$Annotated ``probable inactive'' in
    UniProt. Each joint effect is larger than that of 30 matched sets of distant residues.}
  \label{fig:q-pairs}
\end{figure}

\paragraph{Residues that act together.}
\Cref{fig:q-pairs} compares knocking residues out one at a time with knocking them out together.
For the Cys--His dyad of a deubiquitinase the joint effect exceeds the sum. Two proteins are
annotated as ``probable inactive'' enzymes, yet their catalytic residues are present and the
model's call depends on them jointly: the two aspartates of the protease together cost 3.4 times
the sum of the two alone. In cytoglobin the call rests on the proximal histidine, which binds
the heme iron, and not on the distal one. These are statements about what the model relies on;
whether the two ``inactive'' proteins have activity is a question for experiment.

\begin{figure}[!htbp]
  \centering
  \begin{minipage}[c]{0.40\linewidth}
    \centering
    \includegraphics[width=\linewidth]{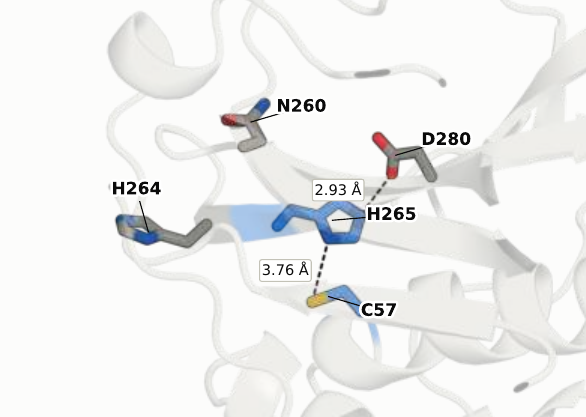}
  \end{minipage}\hfill
  \begin{minipage}[c]{0.57\linewidth}
    \centering
    \begin{tikzpicture}
      \begin{axis}[paper, xbar, bar width=6pt, width=0.70\linewidth, height=5.0cm,
          axis lines*=left, ymajorgrids=false, xmajorgrids, y dir=reverse,
          xmin=-8.6, xmax=0.9, xtick={-6,-4,-2,0}, xlabel={change in the model's function score},
          ytick={1,2,3,4,5,6,7,8.6,9.6},
          yticklabels={Cys57, His265, Asn55, Asp280, Asn260, His264, Trp29, {Cys57 + His265, summed}, {Cys57 + His265, together}},
          y tick label style={font=\scriptsize}, ymin=0.3, ymax=10.3,
          nodes near coords, nodes near coords align={horizontal},
          every node near coord/.append style={font=\tiny, anchor=east, inner sep=2pt,
            /pgf/number format/.cd, fixed, fixed zerofill, precision=2}]
        \fill[cGreyL, opacity=0.5] (axis cs:-0.213,0.3) rectangle (axis cs:0.213,10.3);
        \addplot[fill=cGrey, draw=none, bar shift=0pt] coordinates {(-0.31,2) (-0.18,3) (-0.09,4) (-0.06,5) (0.03,6) (0.04,7)};
        \addplot[reference bar, bar shift=0pt] coordinates {(-5.13,8.6)};
        \addplot[fill=cOurs, draw=none, bar shift=0pt] coordinates {(-4.82,1) (-6.86,9.6)};
      \end{axis}
    \end{tikzpicture}
  \end{minipage}
  \caption{\textbf{Mouse DORIP1 (Q4KL13): a hypothesis, with no experimental validation.}
    \emph{Left:} Cys57 and His265 in the predicted structure (AlphaFold Database), 3.8\,\AA{}
    apart. \emph{Right:} change in the model's function score when a residue is knocked out.
    Removing Cys57 and His265 together costs more than the sum of the two alone; the other
    residues tested fall inside the range of matched controls (grey band).}
  \label{fig:case-dorip}
\end{figure}
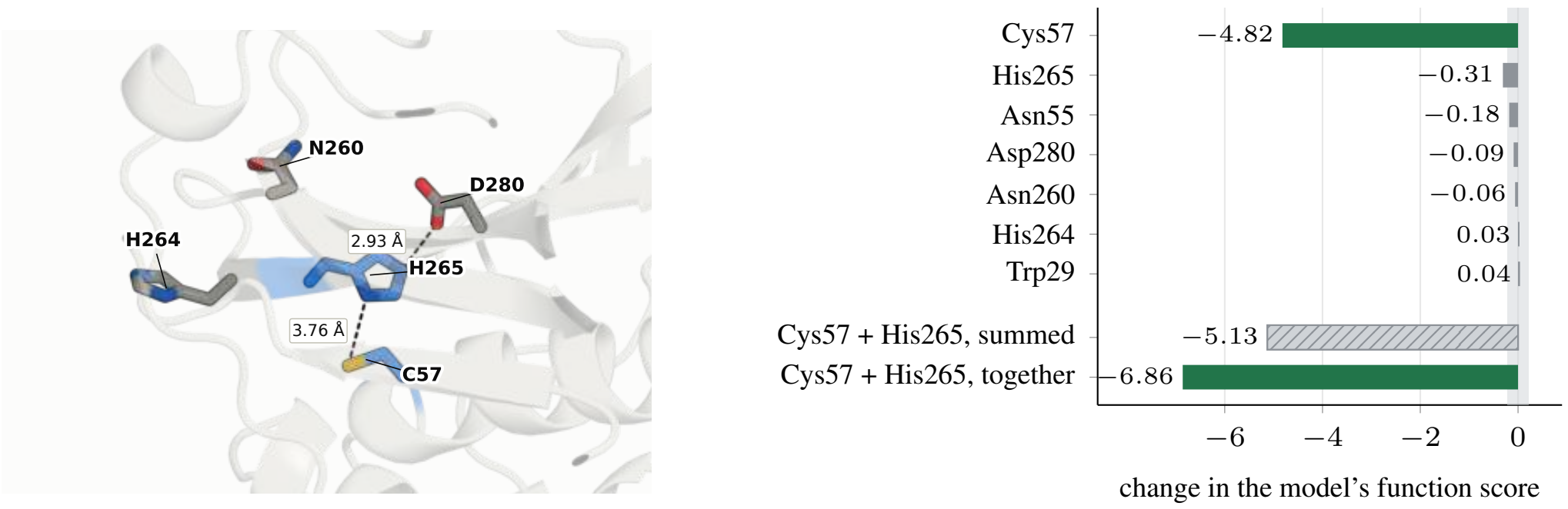

\paragraph{An untested prediction for DORIP1.}
DORIP1 binds the dopamine D1 receptor, but its role as an enzyme has not been established.
Asked about the mouse protein, RumiFold predicted that it could act as a protease, an enzyme that cuts proteins.
This prediction depended most strongly on Cys57 and more weakly on His265.
Although these residues are far apart in the sequence, they lie close together in the predicted structures of both mouse and human DORIP1 (\cref{fig:case-dorip}).

RumiFold's two ways of predicting function gave different answers.
Its written response named a cysteine protease, while its separate function classifier gave low scores to that activity.
Whether DORIP1 actually cuts proteins remains an open question.

\FloatBarrier
\section{Reasoning With Evidence, and Example Answers}
\label{app:qual}

\paragraph{From residues to function, in the model's words.}
During functional training the model is asked to write its answer in a fixed order: the
functional residues it finds, with their distances on its own predicted fold; what they imply;
and then the Gene Ontology lineage from general to specific. \Cref{fig:q-reason} shows four such
answers on held-out proteins. In the first two the residues, the chemistry and the function are
all right, for proteins with no annotated family in training. In the third the model finds the
iron centre and the right class of reaction but names the wrong specific enzyme. In the fourth it
finds the catalytic aspartate and then calls the protein a serine protease: the stated evidence
and the conclusion disagree. Over the 113 proteins of this evaluation the residues and the
function were both right for 57, both wrong for 18, and in 38 one was right without the other.
We showed in \cref{sec:limits} that writing these observations out does not improve function
prediction, and in \cref{app:chain} that the model does not read the rows it cites; the format is
useful as a window on what the model associates, not as evidence of how it computes.

\begin{figure}[H]
\qhead{Zinc metalloendopeptidase (C5FUK3): right}
\qbox{\qlab{Model.} Observations: active site: Glu432. Binds Zn(2+): His431, His435, His441 (C$\alpha$--C$\alpha$ 7--10\,\AA{}, one cluster).
Inference: Cofactor: Zn(2+). The residues form a catalytic site of 4 residues in 2 groups. The chemistry is a hydrolysis.
Function: catalytic activity\gt hydrolase activity\gt peptidase activity\gt endopeptidase activity\gt metalloendopeptidase activity. EC 3.4.24.-.\\
\qref{Reference.} Active site: Glu432, His441. Binds Zn(2+): His431, His435, His441, Cys623. Same chemistry and the same function lineage; EC 3.4.24.-.}
\qhead{Deubiquitinase (Q5VV17): right}
\qbox{\qlab{Model.} Observations: active site: Asp317, Cys320, His431 (C$\alpha$--C$\alpha$ 5--9\,\AA{}, one cluster).
Inference: The residues form a catalytic site of 3 residues. The chemistry is a hydrolysis.
Function: catalytic activity\gt hydrolase activity\gt peptidase activity\gt ubiquitin-like protein peptidase activity\gt deubiquitinase activity\gt cysteine-type deubiquitinase activity. EC 3.4.19.12.\\
\qref{Reference.} The same three residues (5--10\,\AA{}), the same lineage and EC number.}
\qhead{Iron-dependent dioxygenase (D0E8I4): right class, wrong enzyme}
\qbox{\qlab{Model.} Observations: binds Fe cation: His118, Asp120, His198 (C$\alpha$--C$\alpha$ 6--8\,\AA{}, one cluster).
Inference: The residues form an iron centre of 3 residues. The chemistry is an oxidoreduction.
Function: catalytic activity\gt oxidoreductase activity\gt oxidoreductase activity, acting on paired donors, with incorporation or reduction of molecular oxygen\gt 2-oxoglutarate-dependent dioxygenase activity. EC 1.14.11.55.\\
\qref{Reference.} The same residues and lineage, but EC 1.14.11.46, (2-aminoethyl)phosphonate dioxygenase.}
\qhead{Aspartic peptidase (Q45832): the conclusion contradicts the evidence}
\qbox{\qlab{Model.} Observations: active site: Asp174.
Inference: The residues form a catalytic site of 1 residue. The chemistry is a hydrolysis.
Function: catalytic activity\gt hydrolase activity\gt peptidase activity\gt serine-type peptidase activity\gt serine-type endopeptidase activity. EC 3.4.21.-.\\
\qref{Reference.} Active site: Asp174. Function: \ldots\gt endopeptidase activity\gt aspartic-type endopeptidase activity. EC 3.4.23.-.}
\caption{Answers written in the order observations, inference, function, on held-out proteins,
  lightly reformatted (chain labels removed). These come from a checkpoint of the functional
  training stage (step 126k), earlier than the models in the results tables.}
\label{fig:q-reason}
\end{figure}

\begin{table}[h]
  \caption{\textbf{Putting the model's own residue predictions in the prompt lowers function
    accuracy, and their content does not matter.} Information-weighted F1 on the GO-GPT v2 test
    set (3{,}726 proteins), same scorer as \cref{tab:go}. In every row but the first the model
    first states functional residues and then predicts function with that text in its prompt.
    $\Delta$: paired difference in the mean from direct prediction (95\% bootstrap intervals are
    within $\pm 0.008$). The evidence rows and $\Delta$ were measured on an earlier checkpoint,
    whose direct prediction scores 0.682.}
  \label{tab:evidence}
  \centering
  \small
  \begin{tabular}{@{}lccccc@{}}
    \toprule
    Evidence in the prompt & MF & BP & CC & Mean & $\Delta$ \\
    \midrule
    None (function predicted directly)             & \textbf{0.685} & \textbf{0.615} & \textbf{0.748} & \textbf{0.683} & \\
    Own residue predictions                        & 0.656 & 0.495 & 0.722 & 0.624 & $-0.058$ \\
    Own residue predictions, confident ones only   & 0.656 & 0.498 & 0.725 & 0.626 & $-0.056$ \\
    Own free-text observations                     & 0.655 & 0.506 & 0.725 & 0.629 & $-0.053$ \\
    \emph{Controls} & & & & & \\
    \quad Own predictions, positions redrawn at random & 0.657 & 0.497 & 0.717 & 0.623 & $-0.058$ \\
    \quad Another protein's predictions            & 0.655 & 0.499 & 0.722 & 0.625 & \\
    \bottomrule
  \end{tabular}
\end{table}

\paragraph{Does stating the evidence help?}
\Cref{tab:evidence} tests whether the model can use the observations it writes. It cannot. Every
version of the evidence costs 0.053 to 0.058 in mean weighted F1, almost all of it on biological
process ($-0.117$, against $-0.029$ for molecular function and $-0.028$ for cellular component).
The model's own residue predictions do no better than the same statements with the positions
redrawn at random ($+0.001$, $[-0.003, +0.005]$) or than another protein's predictions ($-0.001$,
$[-0.006, +0.004]$): it reacts to the presence of the text and not to what it says. The loss is
largest on proteins with a close relative in training ($-0.083$ at 90\% identity or more, $-0.030$
below 30\%), and no subset we examined shows a reliable gain.

\paragraph{Free-text descriptions of proteins it has not seen described.}
\Cref{fig:q-p2t} shows descriptions written for novel Prot2Text test proteins, whose reference
text does not occur in the training set. The typical error is visible in the middle three: the
model gets the kind of protein and its role right and one specific wrong, a substrate or a
partner. The last is a plain failure.

\begin{figure}[H]
\qhead{P0A625: right}
\qbox{\qlab{Model.} Part of the binding-protein-dependent transport system for molybdenum; probably responsible for the translocation of the substrate across the membrane.\\
\qref{Reference.} Part of the binding-protein-dependent transport system ModABCD for molybdenum; probably responsible for the translocation of the substrate across the membrane.}
\qhead{Q7S133: right role, wrong partner name}
\qbox{\qlab{Model.} Catalytic component of the SWR1 complex which mediates the ATP-dependent exchange of histone H2A for the H2A variant HZT1 leading to transcriptional regulation of selected genes by chromatin remodeling.\\
\qref{Reference.} \ldots exchange of histone H2A for the H2A variant H2A.Z leading to transcriptional regulation of selected genes by chromatin remodeling.}
\qhead{P0AE25: right mechanism, wrong substrate}
\qbox{\qlab{Model.} Uptake of galactose across the boundary membrane with the concomitant transport of protons into the cell (symport system).\\
\qref{Reference.} Uptake of L-arabinose across the cytoplasmic membrane with the concomitant transport of protons into the cell (symport system).}
\qhead{O31266: right enzyme class, wrong substrate}
\qbox{\qlab{Model.} Ring-cleaving dioxygenase involved in oxoquinoline degradation and utilization.\\
\qref{Reference.} Ring-cleaving dioxygenase involved in quinaldine degradation and utilization.}
\qhead{O13563: wrong}
\qbox{\qlab{Model.} Matches the consensus sequence of the alpha chain of parvalbumins.\\
\qref{Reference.} Component of the 19S cap proteasome complex which acts as a regulatory subunit of the 26S proteasome, involved in the ATP-dependent degradation of ubiquitinated proteins.}
\caption{Generated and reference descriptions for five novel Prot2Text test proteins (\ours{} at
  step 240k, greedy decoding).}
\label{fig:q-p2t}
\end{figure}

\paragraph{Asked about itself, the model answers about the protein.}
As an informal probe of how the protein in context colours ordinary conversation, we gave the
Gene Ontology model a protein and asked a question that has nothing to do with it: ``What were
you trained on?'' \Cref{fig:q-trained} shows replies. The answer changes with the protein. With
ACE2, a peptidase, the model answers as if it were the enzyme and lists its substrates; with
insulin it says it was trained as a nurse; with tau and p53 it describes a medical education.
Over 20 proteins, 8 replies describe medical or clinical training, 6 are the stock reply of a
language model, and one, for T4 lysozyme, lists protein sequence and structure data, which is the
nearest to the truth. These are single greedy samples with labels assigned by hand, so they are
anecdotes and not a measured effect. We include them because they show the folding memory
reaching into answers it was never trained to inform, which is one face of the cost to general
language ability reported in \cref{sec:limits}.

\begin{figure}[H]
\qhead{ACE2 (Q9BYF1), a peptidase: answers as the enzyme}
\qbox{\qlab{Model.} I was trained on a variety of substrates, including but not limited to, proteins, peptides, and other dietary materials. My training involved learning how to recognize and respond to different types of stimuli, such to identify a target for processing, such as a peptide or a protein, and then processing it according to the target's requirements.}
\qhead{Insulin (P01308): a clinical persona}
\qbox{\qlab{Model.} I was trained as a nurse.}
\qhead{Tau (P10636): a medical curriculum}
\qbox{\qlab{Model.} I was trained on a comprehensive curriculum that includes foundational knowledge in biology, anatomy, and physiology, followed by specialized instruction in medical sciences, clinical skills, and professional ethics. My training is designed to prepare me for a medical career, with a focus on critical thinking, precision, and patient-centered care.}
\qhead{p53 (P04637): oncology first}
\qbox{\qlab{Model.} I was trained on a vast amount of data, including but not limited to medical literature, research papers, and real-world clinical scenarios. My training is based on information from various domains, such as oncology, cardiology, pharmacology, and more. \ldots}
\qhead{T4 lysozyme (P00720): closest to the truth}
\qbox{\qlab{Model.} I was trained on a large amount of data, including but not limited to: genomic sequences (DNA, RNA, etc.); structural information about proteins and their domains; taxonomic information; functionality (e.g., enzyme substrates, reaction mechanisms); sequence-specific patterns that indicate specific binding or processing. \ldots}
\caption{Replies to ``What were you trained on?'' with a different protein in context, from a
  Gene Ontology fine-tune of \ours{} (greedy decoding; one sample per protein; the T4 lysozyme
  reply is condensed from a list).}
\label{fig:q-trained}
\end{figure}

\end{document}